\documentclass[pdflatex,sn-mathphys,iicol]{sn-jnl}

\usepackage{epsfig}
\usepackage{graphicx}
\usepackage{amsmath,epsfig}
\usepackage{graphicx}
\usepackage{textcomp}
\usepackage{xcolor}
\usepackage{rotating}
\usepackage{multirow}
\usepackage{booktabs}
\usepackage{colortbl}
\usepackage{array}
\usepackage{overpic}

\jyear{2021}%

\def\ie{\emph{~i.e.,~}}
\def\eg{\emph{,~e.g.,~}}

\def\etal{{\em ~et al.~}}
\def\ourmodel {\emph{DFM-Net}}
\definecolor{mygray}{gray}{.92}

\newcommand{\zwb}[1]{{\textcolor{black}{#1}}}

\newcommand{\fkr}[1]{{#1}}

\theoremstyle{thmstyleone}%
\theoremstyle{thmstyletwo}%

\theoremstyle{thmstylethree}%

\begin{document}

\title[Article Title]{Depth Quality-Inspired Feature Manipulation for Efficient RGB-D and Video Salient Object Detection}


\author[1]{\fnm{Wenbo} \sur{Zhang}}\email{zhangwenbo@stu.scu.edu.cn}

\author*[1,2]{\fnm{Keren} \sur{Fu}}\email{fkrsuper@scu.edu.cn}

\author[1]{\fnm{Zhuo} \sur{Wang}}\email{imwangzhuo@stu.scu.edu.cn}

\author[3]{\fnm{Ge-Peng} \sur{Ji}}\email{gepengai.ji@gmail.com}

\author[1,2]{\fnm{Qijun} \sur{Zhao}}\email{qjzhao@scu.edu.cn}


\affil[1]{\orgdiv{College of Computer Science}, \orgname{Sichuan University}}

\affil[2]{\orgdiv{National Key Laboratory of Fundamental Science on Synthetic Vision}, \orgname{Sichuan University}}

\affil[3]{\orgdiv{Institute of Artificial Intelligence, School of Computer Science}, \orgname{Wuhan University}}


\abstract{Recently CNN-based RGB-D salient object detection (SOD) has obtained significant improvement on detection accuracy. However, existing models often fail to perform well in terms of efficiency and accuracy simultaneously. This hinders their potential applications on mobile devices as well as many real-world problems. 
To bridge the accuracy gap between lightweight and large models for RGB-D SOD, in this paper, an efficient module that can greatly improve the accuracy but adds little computation is proposed. Inspired by the fact that depth quality is a key factor influencing the accuracy, we propose an efficient depth quality-inspired feature manipulation (DQFM) process, which can dynamically filter depth features according to depth quality. The proposed DQFM resorts to the alignment of low-level RGB and depth features, as well as holistic attention of the depth stream to explicitly control and enhance cross-modal fusion.   
We embed DQFM to obtain an efficient lightweight RGB-D SOD model called \ourmodel, where we in addition design a tailored depth backbone and a two-stage decoder as basic parts.  
\fkr{
Extensive experimental results on nine RGB-D datasets demonstrate that our \ourmodel~outperforms recent efficient models, running at 20 FPS on CPU with only $\sim$8.5Mb model size, and meanwhile being 2.9/2.4 times faster and 6.7/3.1 times smaller than the latest best models A2dele and MobileSal. It also maintains state-of-the-art accuracy when even compared to non-efficient models. Interestingly, further statistics and analyses verify the ability of DQFM in distinguishing depth maps of various qualities without any quality labels. Last but not least, we further apply \ourmodel~to deal with video SOD (VSOD), achieving comparable performance against recent efficient models while being 3/2.3 times faster/smaller than the prior best in this field. Our code is available at \href{https://github.com/zwbx/DFM-Net}{https://github.com/zwbx/DFM-Net}.
}
\vspace{-0.2cm}}




\maketitle

\section{Introduction}\label{sec:introduction}

Salient object detection (SOD) aims to locate image regions that attract master human visual attention. It is useful in many downstream tasks \eg object segmentation \cite{SaliencyAwareVO}, medical segmentation \cite{fan2020pra,ji2021pnsnet,ji2022vps}, tracking \cite{2019Non}, video object segmentation \cite{ji2021FSNet}, \fkr{and point cloud registration \cite{wan2021rgb}}. Owing to the powerful representation ability of deep learning, great progresses have been made for SOD in recent years, but most of them use only RGB images as input to detect salient objects \cite{RGBsurvey,deng2021re,Jiang2013SalientOD,He2015SuperCNNAS,Tian2022LearningTD,peng2021saliency}. This unavoidably incurs challenges in complex scenarios, such as cluttered or low-contrast background, \fkr{where colors hardly provide clear distinguishable cues.} 

With the popularity of depth sensors/devices, RGB-D SOD has become a hot research topic \cite{BBSNet,fu2021siamese,UCNet-TPAMI,HDFNet,zhang2021bilateral,2020ASIF,2020Going,chen2021cnn}, because additional useful spatial information embedded in depth maps could serve as a complementary cue for more robust detection~\cite{RGBDsurvey,Cong2019Review}. Meanwhile, depth data is already widely available on many mobile devices~\cite{D3Net}\eg{}modern smartphones like Huawei Mate 40 Pro, iPhone 12 Pro, and Samsung Galaxy S20+. This has opened up a new range of applications for the RGB-D SOD task. Unfortunately, the time-space consumption of existing approaches~\cite{JLDCF,D3Net,S2MA,UCNet,SSF,cmMS,DPA,CPFP,DRMA} is still too high, hindering their further applications on mobile devices and real-world problems. Therefore, an efficient and accurate RGB-D SOD model is highly desirable.

\begin{figure}
  \centering
 \centerline{\epsfig{figure=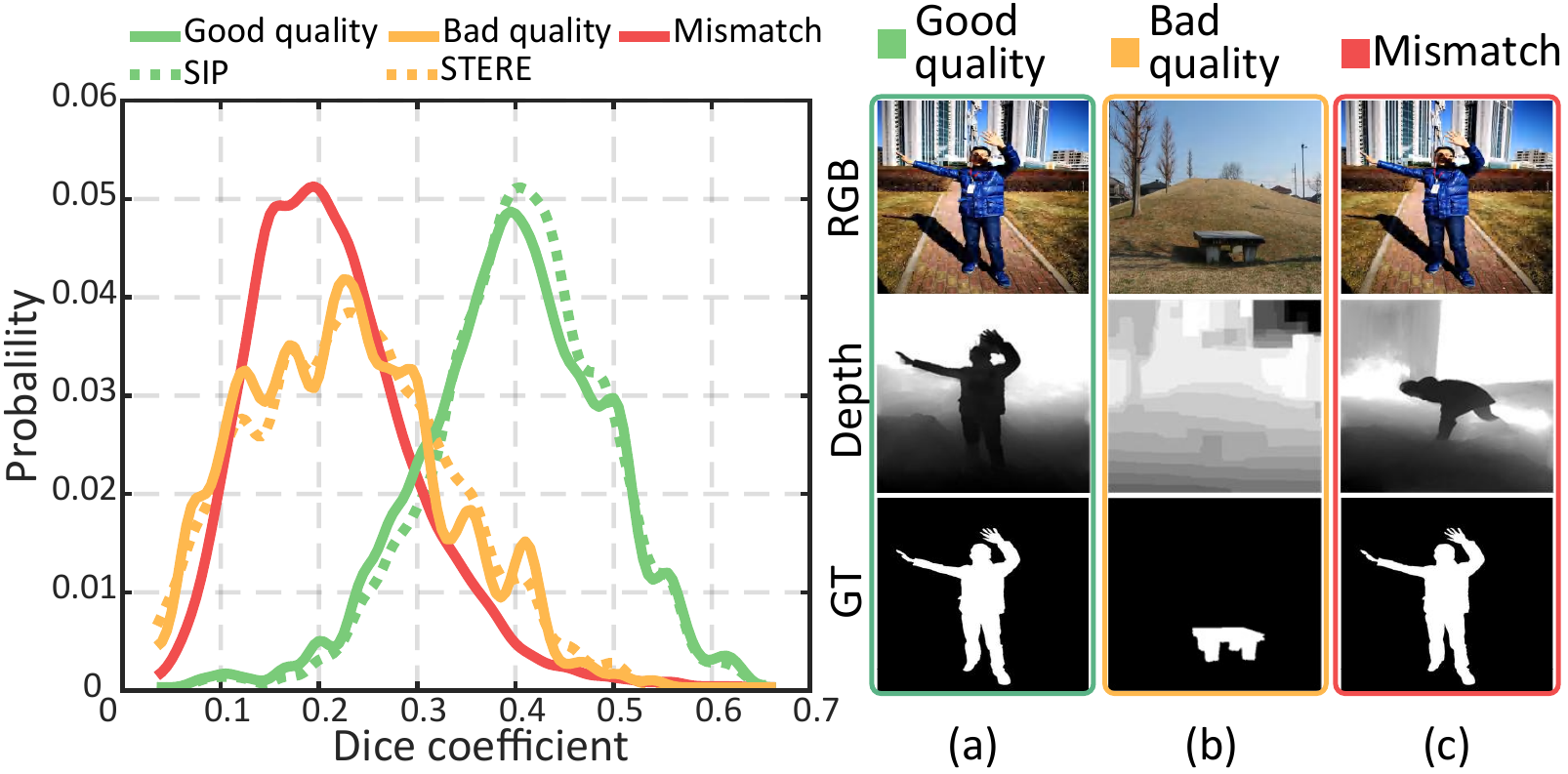,width=0.48\textwidth}}
\caption{Average probability distributions (solid curves) of edge Dice coefficients computed from ``Good quality'', ``Bad quality'', and ``Mismatch'' subsets (10 times random sampling), as well as the probability distributions (dash curves) of the whole STERE and SIP datasets. The right images show three examples from the three subsets, respectively.}
\label{fig:distribution}
\end{figure}

An efficiency-oriented model should aim to significantly decrease the number of operations and memory needed while retaining decent accuracy~\cite{mobilenetV2}. In light of this, several efficiency-oriented RGB-D SOD models have emerged with specific considerations. For example, \cite{A2dele,CoNet,zhao2021rgb} adopt a depth-free inference strategy for fast speed, while depth cues are only utilized in the training phase. \cite{DANet,PGAR,ATSA} choose to design efficient cross-modal fusion modules \cite{DANet} or light depth backbones \cite{PGAR,ATSA}. However, these methods seem to achieve high efficiency at the sacrifice of state-of-the-art (SOTA) accuracy. 
The underlying challenge is that their accuracy is likely to degrade when the corresponding models are simplified or reduced. \fkr{
It is also worth noting that a recent work \cite{wu2021mobilesal}, which is concurrent with ours, proposes a MobileSal (MSal) network that explicitly addresses efficient RGB-D SOD. Although very encouraging performance and speed are reported in \cite{wu2021mobilesal}, to guarantee efficiency, the authors have discarded the popular multi-scale fusion manner and only merge RGB and depth information at the coarsest level, which may result in less sufficient feature utilization and fusion.
}

\textbf{Motivation.} We notice that unstable quality of depth is one key factor which largely influences the accuracy, as mentioned in previous works~\cite{BTSNet,DPA,D3Net,depthconfidence}. However, very few RGB-D SOD models explicitly take this issue into consideration, \fkr{\emph{hence not to mention for efficient models}}~\cite{wu2021mobilesal,PGAR,A2dele}.
We also argue that depth quality is difficult to determine solely according to a depth map itself \cite{depthconfidence,DPA}, because it is tough to judge whether a \fkr{prominent} region in the depth map belongs to noise or a target object, as illustrated in Fig. \ref{fig:distribution} (b). Since RGB-D SOD concerns two paired images as input,\ namely an RGB image and a depth map, we observe that a high-quality depth map usually has some boundaries \emph{well-aligned} to the corresponding RGB image. \fkr{A similar phenomenon is also discussed in \cite{NRDQA} for no-reference depth assessment}. We call this assumption ``boundary alignment'' (BA). 

To validate 
such a BA assumption, we randomly choose 50 paired samples (tagged as ``Good quality'' as shown in Fig. \ref{fig:distribution}) from SIP~\cite{D3Net} dataset, and also 50 ``Bad quality'' samples from STERE \cite{STERE} dataset. \fkr{These two datasets are widely used for RGB-D SOD}, and are chosen here based on the general observations of previous works \cite{SMAC,D3Net,JLDCF}. Additionally, we construct a new set of samples from the ``Good quality'' set, tagged as ``Mismatch'', by randomly mismatching the RGB and depth images of the ``Good quality'' set, to see if this shuffling behavior can be reflected by BA. Note that this behavior actually causes no changes to individual RGB or depth images, therefore having no impact to a depth quality measurement that is dependant only on depth itself (e.g., \cite{depthconfidence}). To quantify boundary alignment, an off-the-shelf edge detector \cite{BDCN} is used to obtain two edge probability maps from RGB and depth, respectively, and then we calculate their Dice coefficient \cite{VNet} ($C_{Dice}$) as a measure of BA:
\begin{gather}
    C_{Dice} = \frac{2  \sum_{i} (E_{RGB}(i) \times E_{Depth}(i))}{\sum_{i} E_{RGB}^2(i) + \sum_{i} E_{Depth}^2(i)},
\end{gather}
where $E_{RGB}$ and $E_{Depth}$ denote RGB and depth edge maps, respectively, and $i$ means the pixel index. $E_{RGB} (i)$/$E_{Depth}(i)$ is the $i$th pixel value on the associated map. \fkr{$C_{Dice}$ will equal to 1 if the two continuous edge maps are perfectly matched.} 
The probability distributions (average of 10 times random sampling) of Dice coefficients are shown in Fig. \ref{fig:distribution}, where the three sets of samples correspond to different colors. We can see that BA seems a strong evidence for the depth quality issue, and meanwhile for the ``Mismatch'' set, its Dice coefficients are generally lower than those of the ``Good quality'' set.

\begin{figure}
\centering
\begin{overpic}[abs,scale=0.8,unit=1mm]{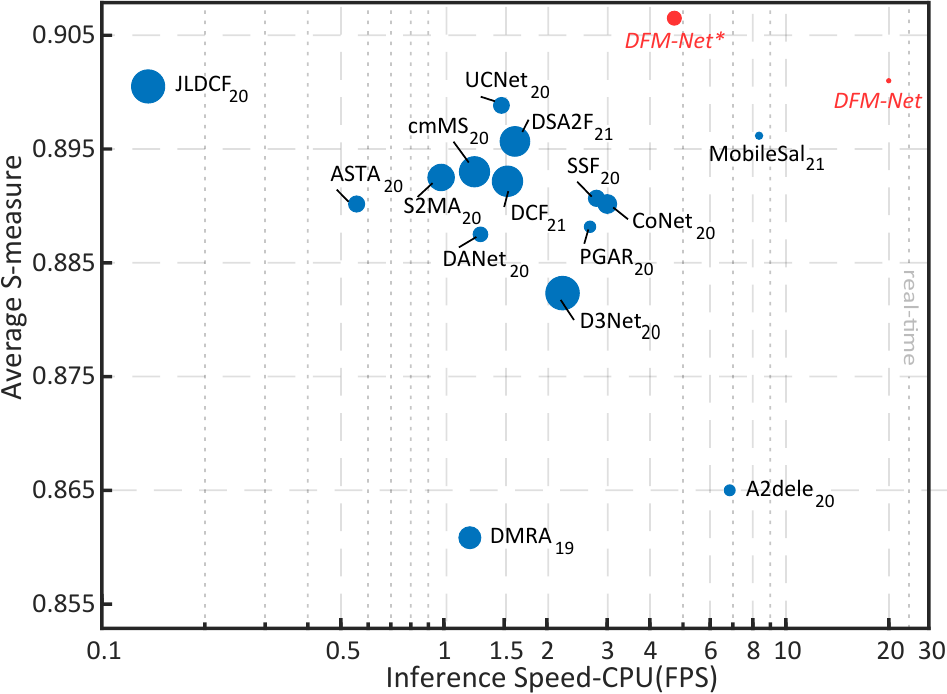}

\put(7,22.3){
    \tiny
	\renewcommand{\arraystretch}{0.05}
	\renewcommand{\tabcolsep}{0.2mm}
	\begin{tabular}{l|c}
	
 Methods&Size(Mb) 
\\
		\specialrule{0em}{1pt}{0pt}
		\hline
		\specialrule{0em}{0pt}{1pt}
D3Net~\cite{D3Net} &530\\JL-DCF~\cite{JLDCF} &520\\DCF~\cite{DCF} &436\\cmMS~\cite{cmMS} &430\\DSA2F~\cite{DSA2F} &415\\S2MA~\cite{S2MA} &330\\DMRA~\cite{DRMA} &228\\CoNet~\cite{CoNet} &167\\SSF~\cite{SSF} &125\\ASTA~\cite{ATSA} &123\\UCNet~\cite{UCNet} &119\\DANet~\cite{DANet} &102\\
\rowcolor{mygray}
\textit{\textbf{DFM-Net*}} &\textbf{93}\\\specialrule{0em}{1pt}{0pt}
		\hline
		\specialrule{0em}{0pt}{1pt} PGAR~\cite{PGAR} &62\\A2dele~\cite{A2dele} &57\\MSal~\cite{wu2021mobilesal} &26\\
\rowcolor{mygray}
\textit{\textbf{DFM-Net}} &\textbf{8.5}\\
		
	\end{tabular}}
\end{overpic}
\vspace{-0.5cm}

\caption{Performance visualization. The vertical axis indicates the average $S_{\alpha}$ \cite{smeasure} over six widely used datasets (NJU2K~\cite{NJU2K}, NLPR~\cite{NLPR}, STERE~\cite{STERE}, RGBD135~\cite{RGBD135}, LFSD\cite{LFSD}, SIP~\cite{D3Net}). The horizontal axis indicates CPU speed (FPS). The circle area is proportional to the model size. More details refer to Tab.~\ref{tab:benchmark}. }
\label{fig:benchmark}
\end{figure} 

Inspired by the above fact that the alignment of low-level RGB and depth features can somewhat reflect depth quality, we propose a new depth quality-inspired feature manipulation (DQFM) process. The intuition behind DQFM is to assign \emph{lower weights} to depth features if the quality of depth is \emph{bad}, effectively avoiding injecting noisy or misleading depth features to improve detection accuracy for efficient models. We also augment DQFM with depth holistic attention, in order to \emph{enhance} depth features when the depth quality is judged to be \emph{good}. With the help of DQFM, we explicitly control and enhance the role of depth features during cross-modal fusion. Finally, we embed DQFM into an encoder-decoder framework to obtain an efficient lightweight model called \ourmodel~\emph{(Depth Feature Manipulation Network)}, where a tailored depth backbone and a two-stage decoder are designed for further efficiency consideration. 
Fig. \ref{fig:benchmark} gives an intuitive performance visualization of the proposed methods \ourmodel~and \ourmodel* (the latter utilizes larger backbones for RGB and depth feature extraction), where 15 SOTA methods and ours are plotted in terms of accuracy, inference speed on CPU, and model size. It is clearly seen that, compared to SOTA techniques, \ourmodel~and \ourmodel* can rank the most upper right with very small circles, indicating that our methods perform better in terms of both efficiency and accuracy. 


In general, our contributions are five-fold:
\vspace{-0.2cm}
\begin{itemize} 
    \item  \fkr{To the best of our knowledge, we are the first to address efficient RGB-D SOD from a depth quality perspective, which is in clear contrast to existing methods~\cite{PGAR,A2dele} as well as the concurrent work MSal \cite{wu2021mobilesal}\footnote{Work \cite{wu2021mobilesal} is concurrent with our preliminary work \cite{zhang2021depth2}.}.}
    
    \item We propose an efficient depth quality-inspired feature manipulation (DQFM) process, to explicitly control and enhance depth features during cross-modal fusion. DQFM avoids injecting noisy or misleading depth features, and can effectively improve detection accuracy with very little time-space addition.
    
    \item Benefited from DQFM, we propose an efficient lightweight model \ourmodel~\emph{(Depth Feature Manipulation Network)},  which has a tailored depth backbone and a two-stage decoder for further efficiency consideration.
      
    \item Compared to 15 SOTA models on nine datasets, \ourmodel~is able to achieve superior accuracy, meanwhile running in quasi real-time at 20 FPS on CPU with only $\sim$8.5Mb model size. Such performance is 2.9$\times$ faster and 6.7$\times$ smaller than the prior best model A2dele~\cite{A2dele}, and also 2.4$\times$ faster and 3.1$\times$ smaller than MSal \cite{wu2021mobilesal}.
    
    \item Concerning video SOD (VSOD) is another saliency field that may demand high efficiency, we apply \ourmodel~to VSOD and achieve quite competitive performance. Meanwhile, we introduce a joint training strategy  for lightweight models to better adapt to VSOD task, which proves to be more effective than the conventional pretraining strategy in this field.
\end{itemize}
    
    
The remainder of this paper is organized as follows. Sec.~\ref{sec:relatedwork} discusses related work on general RGB-D SOD, efficient RGB-D SOD, and depth quality analyses. Sec.~\ref{sec:methodology} describes the proposed \ourmodel~in detail. Experimental results, performance evaluations, and ablation analysis, including the application of \ourmodel~to video SOD, are given in Sec.~\ref{sec:experiment}. Finally, conclusions are drawn in Sec.~\ref{sec:conclusion}. The preliminary version of this work has appeared in \cite{zhang2021depth2}.

\section{Related Work}\label{sec:relatedwork}

The utilization of RGB-D data for SOD has been extensively explored for years. Based on the goal of this paper, in this section, we review general RGB-D SOD methods, as well as previous works on efficient models and depth quality analyses.

\vspace{-0.2cm}

\subsection{General RGB-D SOD Methods}

Traditional methods mainly rely on hand-crafted features \cite{RGBD135,2013An,2012Context,2015Exploiting}. Lately, deep learning-based methods have made great progress and gradually become a mainstream \cite{PCF,HDFNet,UCNet,DRMA,JLDCF,SSF,cmMS,PDNet,CPFP,BBSNet,MMCI,CoNet,D3Net,DANet}. Qu\etal{}\cite{QuRGBD} first introduced CNNs to infer object saliency from RGB-D data.
Zhu\etal{}\cite{PDNet} designed a master network to process RGB data, together with a sub-network for depth data, and then incorporated depth features into the master network. 
Fu \etal{}\cite{JLDCF} utilized a Siamese network for simultaneous RGB and depth feature extraction, which discovers the commonality between these two views from a model-based perspective. Zhang \etal{}\cite{UCNet} proposed a probabilistic network via conditional variational auto-encoders to model human annotation uncertainty. 
Zhang \etal{}\cite{SSF} proposed a complementary interaction fusion framework to locate salient objects with fine edge details. 
Liu \etal{}\cite{S2MA} introduced a selective self-mutual attention mechanism that can fuse attention learned from both modalities.   
Li \etal{}\cite{CMWNet} designed a cross-modal weighting network to encourage cross-modal and cross-scale information fusion from low-, middle- and high-level features. 
Fan \etal{}\cite{BBSNet} adopted a bifurcated backbone strategy to split multi-level features into student and teacher ones, in order to suppress distractors within low-level layers. 
Pang \etal{}\cite{HDFNet} provided a new perspective to utilize depth information, in which the depth and RGB features are combined to generate region-aware dynamic filters to guide the decoding in the RGB stream. 
Li \etal{}\cite{cmMS} proposed a cross-modality feature modulation module that enhances feature representations by taking depth features as prior. 
Luo \etal{}\cite{CAS-GNN} utilized graph-based techniques to design a network architecture for RGB-D SOD. 
Ji \etal{}\cite{CoNet} proposed a novel collaborative learning framework, where multiple supervision signals are employed, yielding a depth-free inference method. 
Zhao \etal{}\cite{DANet} designed a single stream network to directly take a depth map as the fourth channel of an RGB image, and proposed a depth-enhanced dual attention module.

\fkr{
With the flourish in this research direction, other inspiring techniques are also recently employed into the RGB-D SOD task, such as discrepant cross-modality interaction \cite{zhang2021cross}, triplet transformer embedding network \cite{liu2021tritransnet}, pure transformer network \cite{liu2021visual}, neural architecture search \cite{sun2021deep}, mutual information minimization \cite{zhang2021rgb}, specificity-preserving architecture \cite{zhou2021specificity}, hierarchical cross-modal distillation \cite{chen2021cnn}, cross-modal edge-guidance \cite{liu2021cross}, LSTM-based context-aware modules \cite{liang2021context}.
A relatively complete survey on RGB-D SOD can be found in \cite{RGBDsurvey}. 
}

Despite that encouraging detection accuracy has been advanced by the above RGB-D SOD methods, most of them have heavy models and are computationally expensive.

\vspace{-0.2cm}
\subsection{Efficient RGB-D SOD Methods}
Besides the above-mentioned methods, several recent methods attempt to take model efficiency into consideration\footnote{The concept of ``efficient model'' is hard to define. Here and also in the following experiment section, we consider that efficient models should be less than 100Mb.}. 
Specific techniques are used to reduce high computation brought by multi-modal feature extraction and fusion. 
Piao \etal{}\cite{A2dele} employed knowledge distillation for a depth distiller, which aims at transferring depth knowledge obtained from the depth stream to the RGB stream, thus allowing a depth-free inference framework. Chen \etal{}\cite{PGAR} constructed a tailored depth backbone to extract complementary features. Such a backbone is much more compact and efficient than classic backbones\eg{}ResNet \cite{ResNet} and VGGNet \cite{vgg}. Besides, the method adopts a coarse-to-fine prediction strategy that simplifies the top-down refinement process. The utilized refinement module is in a recurrent manner, further reducing model parameters. 
\fkr{More recently, \cite{wu2021mobilesal} introduces an efficient model called MSal for this task, which utilizes an implicit depth restoration technique to strengthen feature representation of the main RGB stream. To ensure high efficiency, depth features are only merged into to the RGB ones at the coarsest level in a small resolution.}

\vspace{-0.2cm}

\subsection{Depth Quality Analyses in RGB-D SOD}
Since the quality of depth often affects model performance, a few researchers have considered the depth quality issue in RGB-D SOD, so as to alleviate the impact of low-quality depth. 
As early attempts, some works proposed to conduct depth quality assessment from a global perspective and obtained a \zwb{quality-inspired score}.
Cong\etal{}~\cite{depthconfidence} first proposed a no-reference depth quality metric \cite{IQA} to alleviate the contamination of low-quality depth. Later, Fan\etal{}~\cite{D3Net} employed triple networks sharing the same structure to process RGB, RGB-D, and depth inputs individually. Depth quality is then evaluated by comparing between the results from the later two sub-networks to explicitly judge whether to introduce depth map information.
Cong\etal{}~\cite{DPA} proposed to predict the depth quality score via a perceptron with high-level RGB and depth features as input. Such a perceptron was trained by scores calculated by comparing thresholded depth maps with ground truth. 
\fkr{Ji\etal{}~\cite{ji2021calibrated} introduced a depth calibration and fusion framework to explicitly address the depth quality issue. It learns to predict a reliability score, based on which the raw depth map is calibrated by the estimated depth map.}

Instead of a single quality-inspired score, more recently, spatial quality evaluation of depth maps was also considered, in order to find valuable depth region.
Wang\etal{}~\cite{DQSF} designed three hand-crafted features to excavate depth following multi-scale methodology. Chen\etal{}~\cite{DQSD} proposed to locate the ``Most Valuable Depth Regions'' of depth by
comparing pseudo ground truth (GT) generated from a sub-network with RGB-D as input, with two saliency maps generated from two sub-networks with RGB/depth as input.

Different from the above existing methods that have high time-space complexity, our depth quality assessment is much more efficient and is more suitable to benefit a lightweight model. Besides, our quality module is end-to-end trainable and is also unsupervised. In contrast, methods \cite{DPA,DQSD,ji2021calibrated} are supervised for quality estimation.
\vspace{-0.2cm}

\begin{figure*}
  \centering
 \centerline{\epsfig{figure=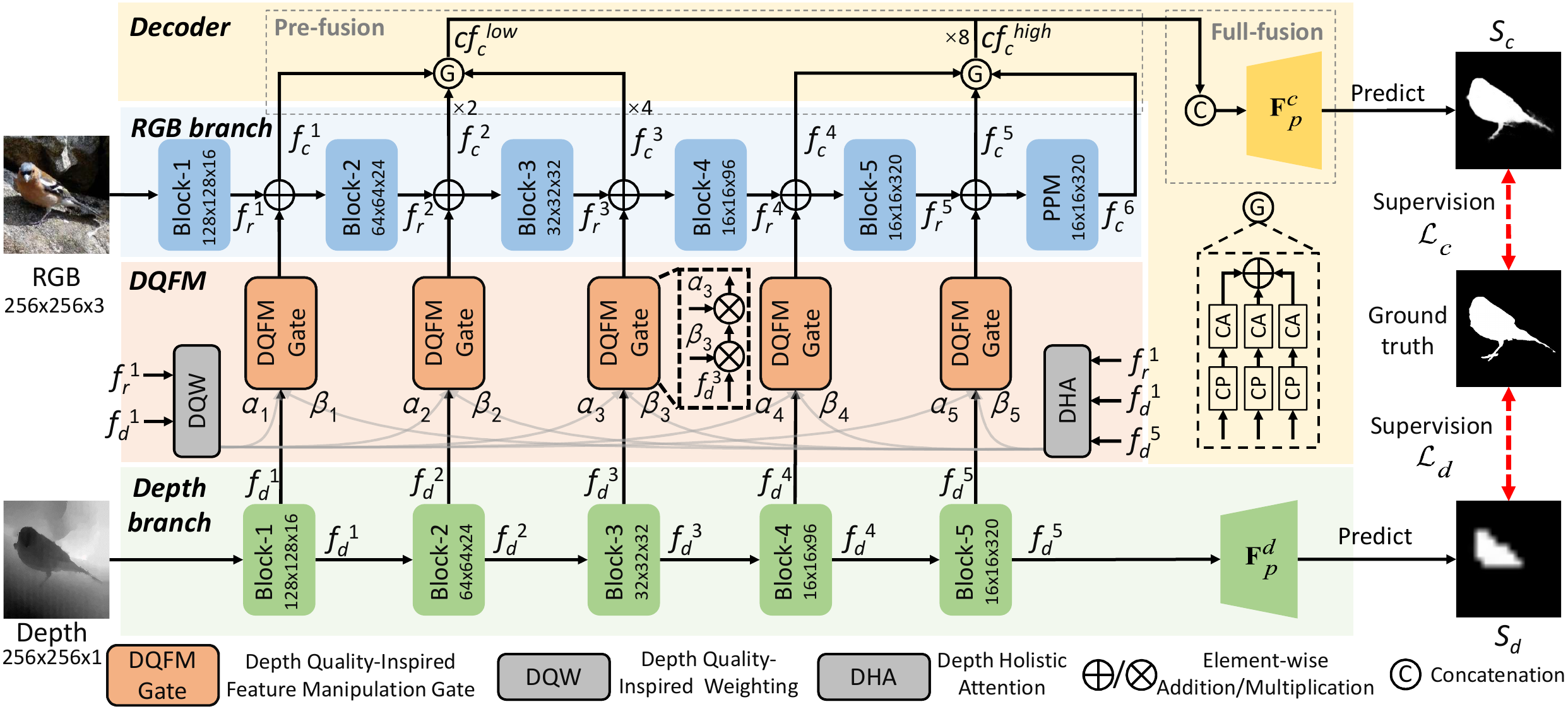,width=1\textwidth}}\vspace{-0.1cm}
\caption{Block diagram of the proposed \ourmodel. Best view in color.}\vspace{-0.1cm}
\label{fig:structure}
\end{figure*}

\section{Methodology}\label{sec:methodology}

\subsection{Overview}

Fig. \ref{fig:structure} shows the block diagram of the proposed \ourmodel, which consists of encoder, decoder, and supervision parts. For efficiency consideration, our encoder part follows the design in \cite{BBSNet}, where the RGB branch is simultaneously responsible for RGB feature extraction and cross-modal fusion between RGB and depth features.
The decoder part, on the other hand, conducts simple two-stage fusion to generate the final saliency map. 
More specifically, the encoder consists of an RGB-related branch which is based on MobileNet-v2~\cite{ResNet}, a depth-related branch which is a tailored efficient backbone, and also the proposed DQFM. Both branches lead to five feature hierarchies, and the output stride is 2 for each hierarchy except 1 for the last hierarchy. The extracted depth features at a certain hierarchy, after passing through a DQFM gate, are then fused into the RGB branch by simple element-wise addition before being sent to the next hierarchy.
Besides, in order to capture multi-scale semantic information, we add a PPM (pyramid pooling module \cite{ppm}) at the end of the RGB branch.
Note that in practice, the DQFM gate contains two successive operations, namely depth quality-inspired weighting (DQW) and depth holistic attention (DHA). 

Let the features from the five RGB/depth hierarchies be denoted as $f_{m}^{i}~(m\in\{r,d\},i=1,...,5)$, the fused features be denoted as $f_{c}^{i}~(i=1,...,5)$, and the features from PPM be denoted as $f_{c}^{6}$. The aforementioned cross-modal feature fusion can be expressed as:
\begin{gather}\label{eq:1}
 f_{c}^{i}= f_{r}^{i} + (\alpha_{i}\otimes\beta_{i}\otimes{f}_{d}^{i}),
\end{gather}
where $\alpha_{i}$ and $\beta_{i}$ are computed by DQW and DHA to manipulate  depth features $f_{d}^{i}$ to be fused, and $\otimes$ denotes element-wise multiplication\footnote{If the multipliers' dimensions are different, before element-wise multiplication, the one with less dimension will be replicated to have the same dimension of the other(s).}. After encoding as shown in Fig. \ref{fig:structure}, $f_{c}^{i}~(i=1,...,5)$ and $f_{c}^{6}$ are fed to the subsequent decoder part. 

\begin{figure}
  \centering
 \centerline{\epsfig{figure=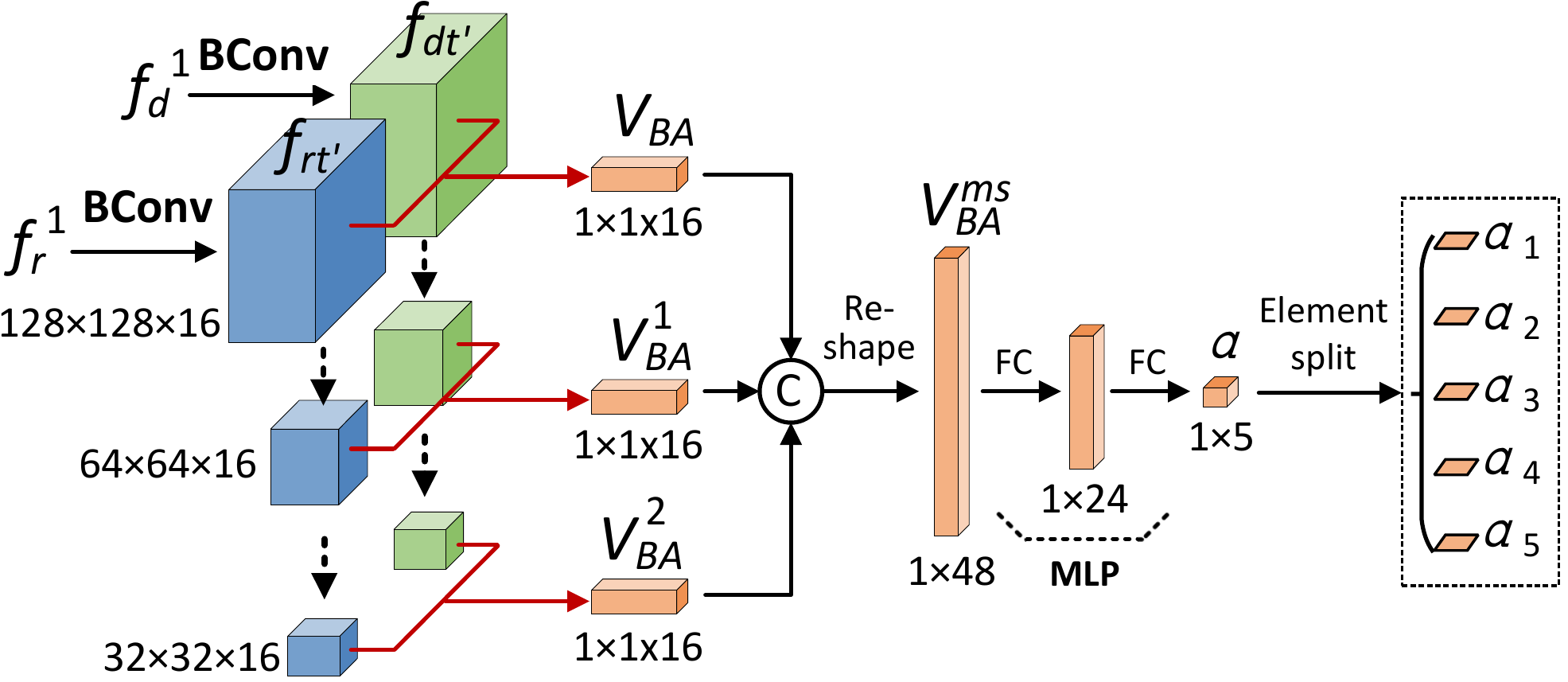,width=0.48\textwidth}}
\caption{Architecture of DQW (depth quality-inspired weighting). The red line arrows indicate the computation of Eq.~\ref{eq:4}. The dash line arrows indicate the max-pooling with stride 2.}
\label{fig:dqw}
\end{figure} 

\subsection{Depth Quality-Inspired Feature Manipulation (DQFM)}\label{sec:DQFM}

DQFM consists of two key components, namely DQW (depth quality-inspired weighting) and DHA (depth holistic attention), which generate $\alpha_{i}$ and $\beta_{i}$ in Eq.  \ref{eq:1}, respectively. $\alpha_{i}\in\mathbb{R}^{1}$ is a scalar determining \emph{``how much''} depth features are involved, while $\beta_{i}\in\mathbb{R}^{k \times k}$ ($k$ is the feature size at hierarchy $i$) is a spatial attention map, deciding \emph{``what regions''} to focus in the depth features. Below, the internal structures of DQW and DHA are introduced.

\subsubsection{Depth Quality-Inspired Weighting (DQW)} Inspired by the aforementioned BA observation in Sec. \ref{sec:introduction}, as shown in Fig.~\ref{fig:dqw}, DQW learns the weighting term $\alpha_{i}$ adaptively from low-level features $f_{r}^{1}$ and $f_{d}^{1}$, because such low-level features characterize image edges/boundaries \cite{Amulet}.
To this end, we first apply convolutions to $f_{r}^{1}$/$f_{d}^{1}$ to obtain transferred features $f_{rt'}$/$f_{dt'}$, which are expected to capture more edge-related activation:
\begin{gather}
f_{rt'}=\textbf{BConv}_{1 \times1}(f_{r}^{1}),~f_{dt'}=\textbf{BConv}_{1 \times1}(f_{d}^{1}),
\end{gather}
where $\textbf{BConv}_{1 \times 1}(\cdot{})$ denotes a $1 \times 1$ convolution with BatchNorm and ReLU activation. To evaluate low-level feature alignment and inspired by the aforementioned Dice coefficient \cite{VNet}, given edge activation $f_{rt'}$ and $f_{dt'}$, the alignment feature vector $V_{BA}$ that encodes the alignment between $f_{rt'}$ and $f_{dt'}$ is computed as:
\begin{gather}
\label{eq:4}
V_{BA}=\frac{\textbf{GAP}(f_{rt'} \otimes f_{dt'})}{\textbf{GAP}(f_{rt'} + f_{dt'})},
\end{gather}
where $\textbf{GAP}(\cdot{})$ denotes the global average pooling operation that aggregates element-wise information, and $\otimes$ means element-wise multiplication.
Here, we empirically drop the square of each element in the denominator of the Dice coefficient, which we find has better performance. Such simplification has also been adopted by some previous works (\emph{e.g.}, \cite{diceloss}).

Further, in order to make $V_{BA}$ robust against slight edge shifting, we compute $V_{BA}$ at multi-scale and concatenate the results to generate an enhanced vector. As shown in Fig.~\ref{fig:dqw}, such multi-scale calculation is conducted by subsequently down-sampling the initial features $f_{rt'}$/$f_{dt'}$ by max-pooling with stride 2, and then computing $V_{BA}^{1},V_{BA}^{2}$ the same way as Eq.~\ref{eq:4}. Suppose $V_{BA}$, $V_{BA}^{1}$, and $V_{BA}^{2}$ are the alignment feature vectors computed from three scales as shown in Fig.~\ref{fig:dqw}, the enhanced vector $V_{BA}^{ms}$ is computed as:
\begin{gather}
V_{BA}^{ms}=[V_{BA},V_{BA}^{1},V_{BA}^{2}],
\end{gather}
where $[\cdot]$ denotes channel concatenation. Next, two fully connected layers are applied to derive $\alpha \in \mathbb{R}^{5}$ from $V_{BA}^{ms}$ as below:
\begin{gather}
\alpha = \textbf{MLP}({V_{BA}^{ms}}),
\end{gather}where $\textbf{MLP}(\cdot{})$ denotes a two-layer perceptron with the Sigmoid function in the end. Thus the vector $\alpha$ obtained contains $\alpha_{i} \in (0,1)~(i={1,2...,5})$ as its elements. Notably, here we adopt different weighting terms for different hierarchies rather than an identical one. The effectiveness of this multi-variable strategy is validated in Sec.~\ref{sec:ablation}.

\begin{figure}
  \centering
 \centerline{\epsfig{figure=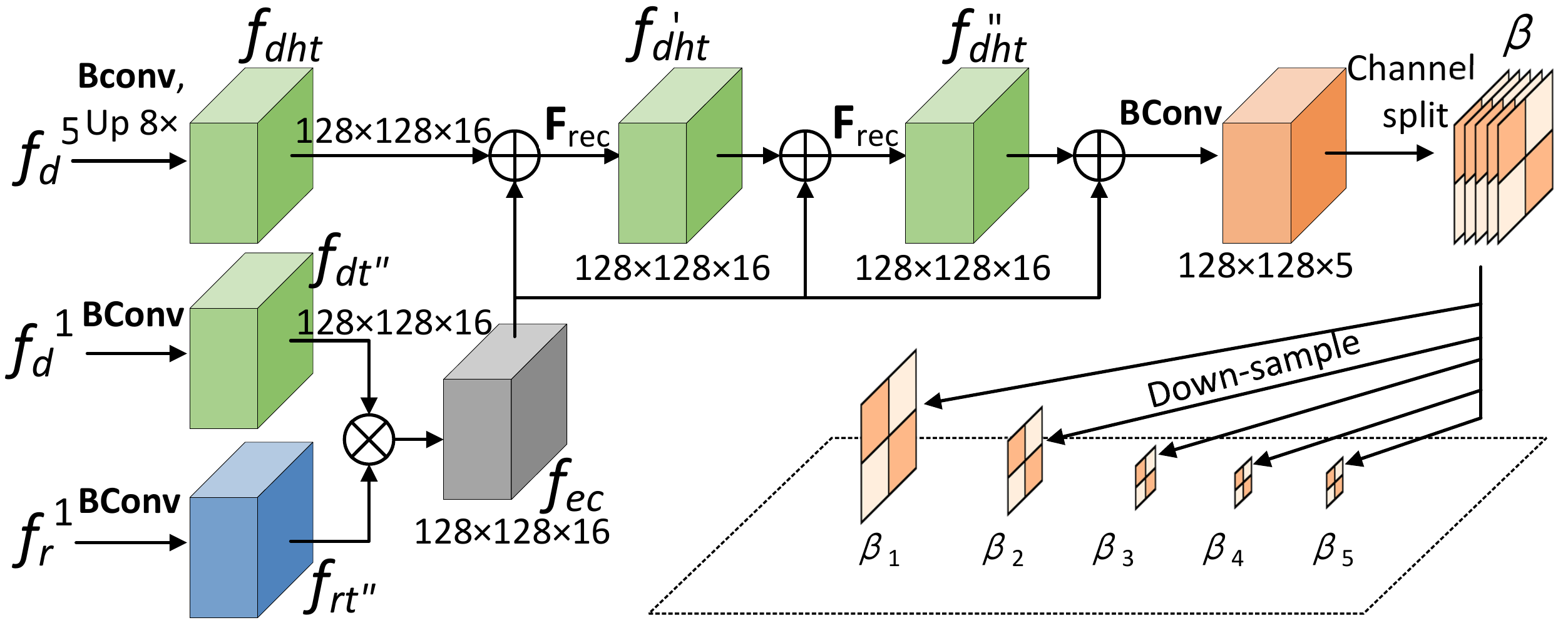,width=0.48\textwidth}}\vspace{0cm}
\caption{Architecture of DHA (depth holistic attention).}\vspace{0cm}
\label{fig:dha}
\end{figure} 
\begin{figure}
  \centering
 \centerline{\epsfig{figure=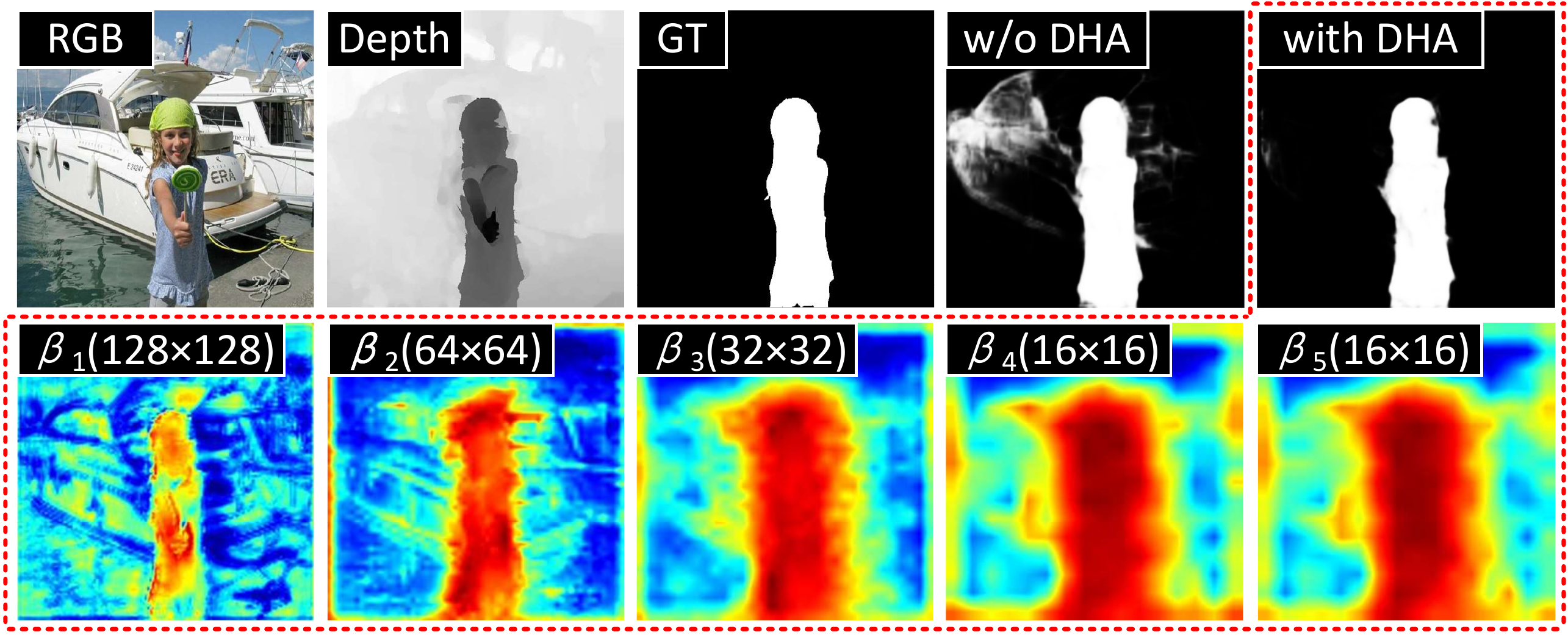,width=0.48\textwidth}}\vspace{-0cm}
\caption{Visualization of holistic attention maps $\beta_1 \sim \beta_5$, as well as the resulting saliency maps with and without DHA.}\vspace{-0.cm}
\label{fig:qualitative_dha}
\end{figure}

\subsubsection{Depth Holistic Attention (DHA)}\label{sec:DHA} Depth holistic attention (DHA) enhances depth features spatially, by deriving holistic attention map $\beta_i$ from the depth stream. Technically as in Fig. \ref{fig:dha}, we first utilize the highest-level features $f_{d}^{5}$ from the depth stream to locate coarse salient regions (with supervision signals imposed as shown in Fig. ~\ref{fig:structure}). To facilitate subsequent pixel-wise operations, we compress and then up-sample $f_{d}^{5}$ into $f_{dht}$, which has the same dimension as $f_{r}^{1}$/$f_{d}^{1}$, formulated as:
\begin{gather}
f_{dht} = \textbf{F}_{UP}^{8}( \textbf{BConv}_{1 \times1}(f_{d}^{5})),
\end{gather}
where $\textbf{F}_{UP}^{8}(\cdot)$ means 8$\times$ bilinear up-sampling. Then we combine low-level RGB and depth features to recalibrate $f_{dht}$. Similar to the computation in DQW, we first transfer $f_{r}^{1}$/$f_{d}^{1}$ to $f_{rt''}$/$f_{dt''}$. The resulting features are element-wisely multiplied to generate features $f_{ec}$, which emphasizes common edge-related activation. To better model long-range dependencies across low-level and high-level features (\ie{}$f_{ec}$ and $f_{dht}$) while maintaining efficiency for DHA, we employ the max-pooling operation and dilated convolution to rapidly increase receptive fields. This recalibration process is defined as:
\begin{gather}
\textbf{F}_{rec}(f_{dht}) = \textbf{F}_{UP}^{2}\Big(\textbf{DConv}_{3 \times 3}\big(\textbf{F}_{DN}^{2}(f_{dht} + f_{ec})\big)\Big),
\end{gather}
where symbol $\textbf{F}_{rec}(\cdot)$ denotes once recalibration process of the input. $\textbf{DConv}_{3 \times 3}(\cdot{})$ denotes the $3 \times 3$ dilated convolution with stride 1 and dilation rate 2, followed by BatchNorm and ReLU activation. $\textbf{F}_{UP}^{2}(\cdot) /  \textbf{F}_{DN}^{2}(\cdot)$ denotes bilinear up-sampling/down-sampling operation to $2/(\frac{1}{2})$ times the original size. As a trade-off between performance and efficiency, we conduct recalibration twice as below:
\begin{gather}
f'_{dht} =\textbf{F}_{rec}(f_{dht}),~f''_{dht} =\textbf{F}_{rec}(f'_{dht}),
\end{gather}
where $f'_{dht}$ and $f''_{dht}$ represent the features recalibrated once and twice, respectively. Finally, we combine $ f''_{dht}$ and the previous $f_{ec}$ to achieve holistic attention maps:
\begin{gather}
    \beta = \textbf{BConv}_{3 \times 3}(f_{ec} + f''_{dht}).
\end{gather}
Note that here the ReLU activation in $\textbf{BConv}_{3 \times 3}$ is replaced by the Sigmoid one, \fkr{in order to achieve attention outputs for $\beta$.} 
Finally, we have obtained five depth holistic attention maps $\{\beta_1,\beta_2,...,\beta_5\}$ by down-sampling from $\beta$, severing as spatial enhancement terms for those depth hierarchies. A visual case of DHA is shown in Fig.~\ref{fig:qualitative_dha}. One can see that $\beta_1$ is likely to highlight regions around edges, while $\beta_5$ focuses more on the dilated entire object. Generally, irrelevant background clutters in depth features can be somewhat suppressed by multiplying attention maps $\beta_1 \sim \beta_5$.

\begin{table}
	\centering
    \scriptsize
	\caption{Detailed structure of the proposed tailored depth backbone (TDB), which is based on inverted residual bottleneck blocks (IRB) of MobileNet-V2 \cite{mobilenetV2}. About notations, $t$:~expansion factor of IRB, $c$: output channels, $n$: times the block is repeated, and $s$: stride of hierarchy, which is applied to the first block of the repeating blocks.}\vspace{-0cm}
	\label{tab:depthbackbone}
	\renewcommand{\arraystretch}{0.7}
	\renewcommand{\tabcolsep}{1mm}
	\begin{tabular}{c|c|c|c|c|c|c}
		\hline\toprule
 Input& Output & Block &$t$ &$c$ &$n$ &$s$
\\
		\midrule
$256\times256\times1$ &$128\times128\times16$ &IRB &3 &16 &1 &2

 \\
	$128\times128\times16$ &$64\times64\times24$ &IRB &3 &24 &3 &2 

  \\
		$64\times64\times24$ &$32\times32\times32$ &IRB &3 &32 &7 &2
\\
$32\times32\times32$ &$16\times16\times96$ &IRB &2 &96 &3 &2\\
$16\times16\times96$ &$16\times16\times320$ &IRB &2 &320 &1 &1\\
		\bottomrule
		\hline
	\end{tabular}
	\vspace{-0.3cm}
\end{table}

\subsection{Tailored  Depth Backbone~(TDB)}\label{sec:DepthBackbone}
Usually, depth is less informative than RGB. Hence we consider using a tailored depth backbone (TDB) to extract depth features, which is much lighter than the off-the-shelf backbone such as MobileNet-V2 but able to maintain comparable accuracy. Specifically, we base our TDB on the inverted residual bottleneck blocks (IRB \cite{mobilenetV2}), and construct a new smaller backbone with reduced channel numbers and stacked blocks.
\fkr{
IRB is the basic block of MobileNet-V2, which distinguishes it from previous MobileNet \cite{mobilenet} and ResNet \cite{ResNet}. In contrast to the ResNet block, the intermediate layers of IRB are designed to be wider than its input/output, so that the inverted residual just connects to the thin bottleneck. Besides, similar to MobileNet, IRB also utilizes depth-separable convolution, but removes the last bottleneck non-linearity for improved performance. The detailed structure of IRB refers to \cite{mobilenetV2}.}   

\fkr{Based on IRB, the structure of TDB is detailed in Tab.~\ref{tab:depthbackbone}. While output channels $c$ and hierarchy stride $s$ are made aligned to the RGB branch, $t$ and $n$ are empirically designed inspired by \cite{mobilenetV2}. More specifically, we reduce the $t$ of MobileNet-V2 by nearly half, and $n$ is determined by referring to the corresponding layers of MobileNet-V2 (those with the same output channels), meanwhile merging channels of certain adjacent layers to maintain the overall bottleneck number (Ours: 15 \emph{vs.} MobileNet-V2's: 17). This finally results in a \{1,3,7,3,1\} aligned sequential design of TDB.}
As a result, our TDB is much lighter than previous lightweight backbones (Ours: only 0.9Mb, ATSA's \cite{ATSA}: 6.7Mb, PGAR's~\cite{PGAR}: 6.2Mb, MobileNet-V2: 6.9Mb), and meanwhile, its performance is competitive and even slightly better than MobileNet-V2 in our \ourmodel~implementation (see Sec.~\ref{sec:ablation}). 

During training, we embed TDB into \ourmodel~with random initialization, and supervision signals are imposed at the end of the backbone to enforce saliency feature learning from depth, as shown in Fig. \ref{fig:structure}. The coarse prediction result obtained from TDB is formulated as: 
\begin{gather}
    S_d = \textbf{F}_p^{d}(f_{d}^{5}),
\end{gather}
where $S_d$ means the coarse prediction from TDB, which is supervised by ground truth (GT). $\textbf{F}_p^{d}(\cdot)$ denotes a prediction head consisting of a $1 \times 1$ convolution followed by Sigmoid activation, and also $16 \times$ bilinear up-sampling to recover the original input size. The effectiveness of the proposed TDB will be validated in Sec.~\ref{sec:ablation} .

\subsection{Two-Stage Decoder}\label{sec:Decoder}
Unlike the popular U-Net \cite{unet} which adopts the hierarchy-by-hierarchy top-down decoding strategy, we propose a simplified two-stage decoder, including pre-fusion and full-fusion, to further improve efficiency. The pre-fusion aims to reduce feature channels and hierarchies, by channel compression and hierarchy grouping, denoted as ``CP'' and ``G'' in Fig. \ref{fig:structure}. Based on the outputs of pre-fusion, the full-fusion further aggregates low-level and high-level hierarchies to generate the final saliency map. \fkr{It is also worth noting that in our decoder, depth-wise separable convolutions are mainly adopted for large numbers of input channels, instead of ordinary convolutions.}

\subsubsection{Pre-fusion Stage} We first use $3\times3$ depth-wise separable convolution \cite{mobilenet} with BatchNorm and ReLU activation, denoted as $\textbf{DSConv}_{3 \times 3}(\cdot)$, to compress the encoder features ($f_{c}^{i},i=1,2...6$) to a unified channel 16. Such an operation is denoted as ``CP'' in Fig. \ref{fig:structure}. Then we use the well-known channel attention operator \cite{senet} $\textbf{F}_{CA}(\cdot)$ to enhance features by weighting different channels, denoted as ``CA'' in Fig.~\ref{fig:structure}. The above process can be described as:
\begin{gather}
    cf_{c}^{i}=\textbf{F}_{CA}(\textbf{DSConv}_{3 \times 3}(f_{c}^{i})),
\end{gather}
where $cf_{c}^{i}$ denotes the compressed and enhanced features.
To reduce feature hierarchies, inspired by \cite{BBSNet}, we group 6 hierarchies into two, namely the low-level hierarchy and the high-level hierarchy, as below:
\begin{gather}
    cf_{c}^{low}=\sum_{i=1}^{3}\textbf{F}_{UP}^{2^{i-1}}(cf_{c}^{i}),~cf_{c}^{high}=\sum_{i=4}^{6}cf_{c}^{i},
\end{gather}
 where $\textbf{F}_{UP}^{i}$ is bilinear up-sampling to $i$ times the original size.
 
\subsubsection{Full-fusion Stage} Since in the pre-fusion stage, the channel numbers and hierarchies are already reduced, in the full-fusion stage, we directly concatenate high-level and low-level hierarchies, and then feed the concatenation to a prediction head to achieve the final  full-resolution prediction map, denoted as:
\begin{gather}
    S_c = \textbf{F}_p^{c}\bigg(\big[cf_{c}^{low},\textbf{F}_{UP}^{8}(cf_{c}^{high})\big]\bigg),
\end{gather}
where $S_c$ is the final saliency map, and $\textbf{F}_{p}^{c}(\cdot)$ indicates a prediction head consisting of two $3\times 3$ depth-wise separable convolutions (followed by BatchNorm layers and ReLU activation), a $3\times 3$ convolution with Sigmoid activation, as well as a $2\times$ bilinear up-sampling to recover the original input size. 

\subsection{Loss Function}\label{sec:loss}
The overall loss $\mathcal{L}$ is composed of the final loss $\mathcal{L}_{c}$ and deep supervision for the depth branch loss $\mathcal{L}_{d}$, formulated as:
\begin{gather}
\mathcal{L}=\mathcal{L}_{c}(S_c,G)+\mathcal{L}_{d}(S_d,G),
\end{gather}
where $G$ denotes the ground truth (GT). Similar to previous works \cite{JLDCF,BBSNet,HDFNet,D3Net}, we use the standard cross-entropy loss for $\mathcal{L}_{c}$ and $\mathcal{L}_{d}$. 


\section{Experiments and Results}\label{sec:experiment}

\begin{table*}[htbp]
	\centering
	\caption{Quantitative benchmark results. The model size, $T_{CPU}$, $S_{GPU}$, and $S_{GPU}^*$ are measured by Mb/ms/FPS/FPS respectively. $\uparrow$/$\downarrow$ for a metric denotes that a larger/smaller value is better. Our results are highlighted in \textbf{bold}. The scores/numbers better than ours are \underline{underlined} (efficient and non-efficient models are labeled separately).}\vspace{-0cm}
	\label{tab:benchmark}
    \tiny
	\renewcommand{\arraystretch}{1.2}
	\renewcommand{\tabcolsep}{0.3mm}
	\begin{tabular}{lr|cccccccccccc|c||ccc|c}
		\hline\toprule
		&\multirow{4}{*}{Metric}\centering      & DMRA &D3Net   &JL-DCF &UCNet  &SSF  &S2MA &CoNet  &cmMS &DANet &ATSA &DCF &DSA2F  &\textbf{\ourmodel$^{*}$} &A2dele &PGAR &  MSal &\textbf{\ourmodel}  \\&   &\tiny ICCV &\tiny TNNLS   &\tiny CVPR &\tiny CVPR   &\tiny CVPR
		&\tiny CVPR
		&\tiny ECCV  &\tiny ECCV
		&\tiny ECCV &\tiny ECCV &\tiny CVPR &\tiny CVPR & \tiny \textbf{Ours} &\tiny CVPR &\tiny ECCV &  TPAMI &\tiny \textbf{Ours}\\ &     &\tiny 2019 &\tiny 2020   &\tiny 2020 &\tiny 2020   &\tiny 2020
		&\tiny 2020
		&\tiny 2020   &\tiny 2020
		&\tiny 2020 &\tiny 2020 	&\tiny 2021 &\tiny 2021 & \tiny \textbf{-} &\tiny2020&\tiny 2020 &  2021 &\tiny \textbf{-} \\
		&     &\cite{DRMA} &\cite{D3Net}   &\cite{JLDCF} &\cite{UCNet}  &\cite{SSF} &\cite{S2MA} &\cite{CoNet}	&\cite{cmMS} &\cite{DANet}  &\cite{ATSA} &\cite{DCF} &\cite{DSA2F}  &-&\cite{A2dele} &\cite{PGAR}&  \cite{wu2021mobilesal}&\textbf{-} \\
		\specialrule{0em}{1pt}{0pt}
		\hline\hline
		\specialrule{0em}{0pt}{1pt}

\multirow{3}{*}{\begin{sideways}\textit{}\end{sideways}}

&Model size (Mb)~$\downarrow$		&228	&530	&520	&119		&125	&330	&167		&430	&102	&123 &436	&415  &\textbf{93} &57 &62 &  26	&\textbf{8.5}
\\

&$T_{CPU}$ (ms)~$\downarrow$  
&840	&450	&7326	&679	&358	&1019	&333	&814	&782	&1800 &652	&620	&\textbf{150} &146 	&374	&120	&\textbf{50}

 \\
		&$S_{GPU}$ (FPS)~$\uparrow$  
		&15	&50	&12	&48	&28	&25	&59	&19	&47	&32 &21	&19	 &\textbf{64} &\underline{184} 	&48	&65	&\textbf{70}

\\
		&$S_{GPU}^*$ (FPS)~$\uparrow$ 	&15	&69	&17	&129	&100	&31	&108	&27	&57	&46 &51	&25	&\textbf{180} &225 	&84	&260	&\textbf{370}

 \\
	    \midrule
		\multirow{4}{*}{\begin{sideways}\textit{SIP}\end{sideways}}
		& $S_{\alpha}\uparrow$ 		&.806	&.860	&.879 &.875		&.874	&.878	&.858		&.867	&.878	&.864 &.876	&.862  &\textbf{.885} &.829	&.875 &  .873 &\textbf{.883}
\\
		& $F_{\beta}^{\rm m}\uparrow$  		&.821	&.861	&.885 &.879		&.880	&.884	&.867		&.871	&.884	&.873 &.884	&.875 &\textbf{.890} &.834	&.877 &.883 &\textbf{.887}
 \\
		& $E_{\xi}^{\rm m}\uparrow$     		&.875	&.909	&.923 &.919		&.921	&.920	&.913		&.091	&.920	&.911 &.922	&.912   &\textbf{.926} &.889	&.914&  .920&\textbf{.926}
  \\
		& $\mathcal{M}\downarrow$ 	&.085	&.063	&.051	&.051	&.053	&.054	&.063	&.061	&.054	&.058 &.052	&.057  &\textbf{.049} &.070 &.059&  .053&\textbf{.051}
 \\
		\midrule
		\multirow{4}{*}{\begin{sideways}\textit{NLPR}\end{sideways}}
		& $S_{\alpha}\uparrow$ 		&.899	&.912	&.925	&.920		&.914	&.915	&.908		&.915	&.915	&.907 &.924	&.919  &\textbf{.926} &.890 &.918&  .920	&\textbf{.923}
 \\
		& $F_{\beta}^{\rm m}\uparrow$    		&.879	&.897	&\underline{.916} &.903	&.896	&.902	&.887		&.896	&.903	&.876 &.912	&.906  &\textbf{.912} &.875	&.898 &  .908&\textbf{.908}
 \\
		& $E_{\xi}^{\rm m}\uparrow$   		&.947	&.953	&\underline{.962} &.956		&.953	&.950	&.945		&.949	&.953	&.945 &.963	&.952  &\textbf{.961} &.937 &.948&  \underline{.961}	&\textbf{.957}
 \\
		& $\mathcal{M}\downarrow$ 		&.031	&.025	&\underline{.022}	&.025	&.026	&.030  &.031		&.027	&.029	&.028 &\underline{.022}	&.024  &\textbf{.024}	&.031&.028&  \underline{.025}&\textbf{.026}
 \\
		\midrule
		\multirow{4}{*}{\begin{sideways}\textit{NJU2K}\end{sideways}}
		& $S_{\alpha}\uparrow$		&.886	&.900	&.903	&.897		&.899	&.894	&.895		&.900	&.891	&.901 &.904	&.895  &\textbf{.912}	&.868&.906&  .905 &\textbf{.906}
 \\
		& $F_{\beta}^{\rm m}\uparrow$  		&.886	&.900	&.903	&.895		&.896	&.889	&.892	 &.897 &.880	&.893  &.906	&.897   &\textbf{.913} &.872 &.905&  .905	&\textbf{.910}
 \\
		& $E_{\xi}^{\rm m}\uparrow$		&.927	&.950	&.944	&.936		&.935	&.930	&.937		&.936	&.932	&.921 &.950	&.936  &\textbf{.950} &.914&.940&  .942	&\textbf{.947}
 \\
		& $\mathcal{M}\downarrow$   		&.051	&.041	&.043	&.043		&.043	&.053	&.047		&.044	&.048 	&.040 &.040 	&.044 &\textbf{.039} &.052&.045&  \underline{.041} &\textbf{.042}
 \\
		\midrule
		\multirow{4}{*}{\begin{sideways}\textit{RGBD135}\end{sideways}}
		& $S_{\alpha}\uparrow$  	&.900	&.898	&.929	&.934		&.905	&\underline{.941}	&.910		&.932	&.904	&.907 &.905	&.917	 &\textbf{.938} &.884&.894&  .929&\textbf{.931}
  \\
		& $F_{\beta}^{\rm m}\uparrow$  		&.888	&.885	&.919	&.930		&.883	&\underline{.935}	&.896		&.922	&.894	&.885 	&.894	&.916 &\textbf{.932} &.873 &.879&  \underline{.924}	&\textbf{.922}
  \\
		& $E_{\xi}^{\rm m}\uparrow$ 		&.943	&.946	&.968	&\underline{.976}		&.941	&{.973}	&.945		&.970	&.957	&.952 &.951	&.954  &\textbf{.973}	&.920&.929&  .970 &\textbf{.972}
\\
		& $\mathcal{M}\downarrow$ 		&.030	&.031	&.022	& .019		&.025	&.021	&.029		&.020	&.029	&.024 &.024	&.023  &\textbf{.019} &.030&.032&  .021	&\textbf{.021}
 \\
		\midrule
		\multirow{4}{*}{\begin{sideways}\textit{LFSD}\end{sideways}}
		& $S_{\alpha}\uparrow$ 		&.839	&.825	&.862	&.864	&.859	&.837	&.862	&.849	&.845	&.865 &.842	&\underline{.883}
		 &\textbf{.870} &.834 & .833 &  .847&\textbf{.865}
\\
		& $F_{\beta}^{\rm m}\uparrow$   		&.852	&.810	&.866	&.864		&\underline{.867} &.835	&.859		&\underline{.869}	&.846	&.862 &.842	&\underline{.889}  &\textbf{.866} &.832&.831&  .841&\textbf{.864}
\\
		& $E_{\xi}^{\rm m}\uparrow$  		&.893	&.862	&.901	&\underline{.905}		&.900 &.873	&\underline{.907}		&.896	&.886	&\underline{.905}  &.883	&\underline{.924} 
		&\textbf{.903} &.874 &.893&  .888&\textbf{.903}
\\
		& $\mathcal{M}\downarrow$  	&.083	&.095	&.071	&\underline{.066}	&\underline{.066}	&.094	&.071		&.074	&.083	&\underline{.064} &.075	&\underline{.055}  &\textbf{.068}	&.077&.093&  .078 &\textbf{.072}
\\
		\midrule
		\multirow{4}{*}{\begin{sideways}\textit{STERE}\end{sideways}}
		&$S_{\alpha}\uparrow$  	&.835	&.899	&.905	&.903		&.893	&.890	&.908		&.895	&.892	&.897 &.902	&.898	 &\textbf{.908}&.885 &\underline{.903}&  \underline{.903}&\textbf{.898}
 \\
		& $F_{\beta}^{\rm m}\uparrow$ 		&.847	&.891	&.901	&.899		&.890	&.882	&.904	&.891	&.881	&.884	&.901	&.900  &\textbf{.904}	&.885&.893&  \underline{.895}&\textbf{.893}
 \\
		& $E_{\xi}^{\rm m}\uparrow$  		&.911	&.938	&.946	&.944		&.936	&.932	&.948		&.937	&.930	&.921 &.945	&.942  &\textbf{.948}	&.935&.936&  .940 &\textbf{.941}
\\
		& $\mathcal{M}\downarrow$   		&.066	&.046	&.042	&\underline{.039}		&.044	&.051	&.040		&.042	&.048	&\underline{.039} &\underline{.039}	&\underline{.039} 	&\textbf{.040} &.043 &\underline{.044}&  \underline{.041}& \textbf{.045}
\\

	\midrule
		\multirow{4}{*}{\begin{sideways}\textit{ReDWeb-S}\end{sideways}}
		
&$S_{\alpha}\uparrow$	&.592	&.689	&\underline{.734}	&.713	&.595	&.711	&.696	&.656	&.693	&.679	&.709	&N/A	&\textbf{.715}	&.641	&.656	&.652	&\textbf{.705}	\\

& $F_{\beta}^{\rm m}\uparrow$	&.579	&.673	&\underline{.727}	&\underline{.710}	&.558	&.696	&.693	&.632	&.679	&.673	&\underline{.714}	&N/A	&\textbf{.700}	&.603	&.632	&.610	&\textbf{.686}	\\

& $E_{\xi}^{\rm m}\uparrow$ 	&.721	&.768	&\underline{.805}	&\underline{.794}	&.710	&.781	&.782	&.749	&.778	&.758	&\underline{.789}	&N/A	&\textbf{.788}	&.672	&.749	&.722	&\textbf{.783}	\\

& $\mathcal{M}\downarrow$	&.188	&.149	&\underline{.128}	&\underline{.130}	&.189	&.139	&.147	&.161	&.142	&.155	&.135	&N/A	&\textbf{.132}	&.160	&.161	&.162	&\textbf{.136}

\\

		\bottomrule
		\hline
	\end{tabular}
\begin{flushleft}
	{Note the speed test results here differ from those in our preliminary version \cite{zhang2021depth2}, because we later switched from Windows 10 to Ubuntu 16.04 and re-evaluated all models. We found the latter was more compatible with model implementations, allowing us to better standardize the test environment. We also found that generally the inference speed on Ubuntu was faster.}
\end{flushleft}	
\vspace{-0.cm}
\end{table*}

\subsection{Datasets and Metrics}
We conduct experiments on nine public datasets, including NJU2K \cite{NJU2K} (1,985 samples), NLPR \cite{NLPR} (1,000 samples), STERE \cite{STERE} (1,000 samples), RGBD135 \cite{RGBD135} (135 samples), LFSD \cite{LFSD} (100 samples), SIP \cite{D3Net} (929 samples), \zwb{DUT-RGBD~\cite{DRMA} (1,200 samples), RedWeb-S~\cite{S2MA} (1,000 samples, only the testing set) as well as COME~\cite{zhang2021rgb} (15,625 samples)}. 

Four commonly used metrics are used for evaluation, including S-measure ($S_{\alpha}$) \cite{smeasure},  maximum F-measure ($F_{\beta}^{\rm m}$) \cite{fmeasure}, maximum E-measure ($E_{\xi}^{\rm m}$) \cite{emeasure,Fan2018Enhanced}, and mean absolute error (MAE, $\mathcal{M}$) \cite{mae}. For these four metrics, higher $S_{\alpha}$, $F_{\beta}^{\rm m}$, $E_{\xi}^{\rm m}$, and lower $\mathcal{M}$ indicate better performance.
For efficiency analysis, we report the model size (Mb, Mega-bytes), inference time $T_{CPU}$ (ms, millisecond) on CPU, and one-/maximum-batch inference speed $S_{GPU}/S_{GPU}^*$ (FPS, frame-per-second) on GPU. The maximum-batch setting is previously adopted by Wu\etal{}~\cite{wu2021mobilesal} because it would better reflect the efficiency superiority of light-weight models\footnote{This is because depth-wise separable convolutions-based lightweight models often reach a bottleneck in read/write rate first prior to reaching maximal GPU utilization \cite{Rooflinemodel}, so their acceleration on GPU and efficiency advantages cannot be fully reflected in the small batch setting.}. In the following, we illustrate how $S_{GPU}$ and $S_{GPU}^*$ are measured. Let the inference speed $S$ be defined as:
\begin{gather}
    S = \frac{N \times B}{T},
\end{gather}
where $N$ means the number of total inference, $B$ denotes the batch size used, and $T$ is the total consumed time (in seconds). $N$ is empirically set to 100 in our evaluation, which is large enough for time statistics. When $B$ equals to 1, $S$ becomes $S_{GPU}$, and when $B$ increases to reach the maximal (100\%) GPU utilization, $S$ then becomes $S_{GPU}^*$.

\vspace{-0.2cm}
\subsection{Implementation Details}\label{sec:implementation_details}

Experiments are conducted on a workstation with Intel Core i7-8700 CPU, NVIDIA GTX 1080 Ti GPU. We implement \ourmodel~by Pytorch \cite{pytorch}, and RGB and depth images are both resized to $256 \times 256$ for input. 
For training, in order to generalize the network on limited training samples, following \cite{BBSNet}, we apply various data augmentation techniques \ie{}random translation/cropping, horizontal flipping, color enhancement and so on. We train \ourmodel~for 300 epochs on a single 1080 Ti GPU, taking about 4 hours. The initial learning rate is set as 1e-4 for Adam optimizer~\cite{kingma2014adam}, and the batch size is 10. The poly learning rate policy is used, where the power is set to 0.9.
 \vspace{-0.2cm}
 
\begin{figure*}[htbp]
  \centering
 \centerline{\epsfig{figure=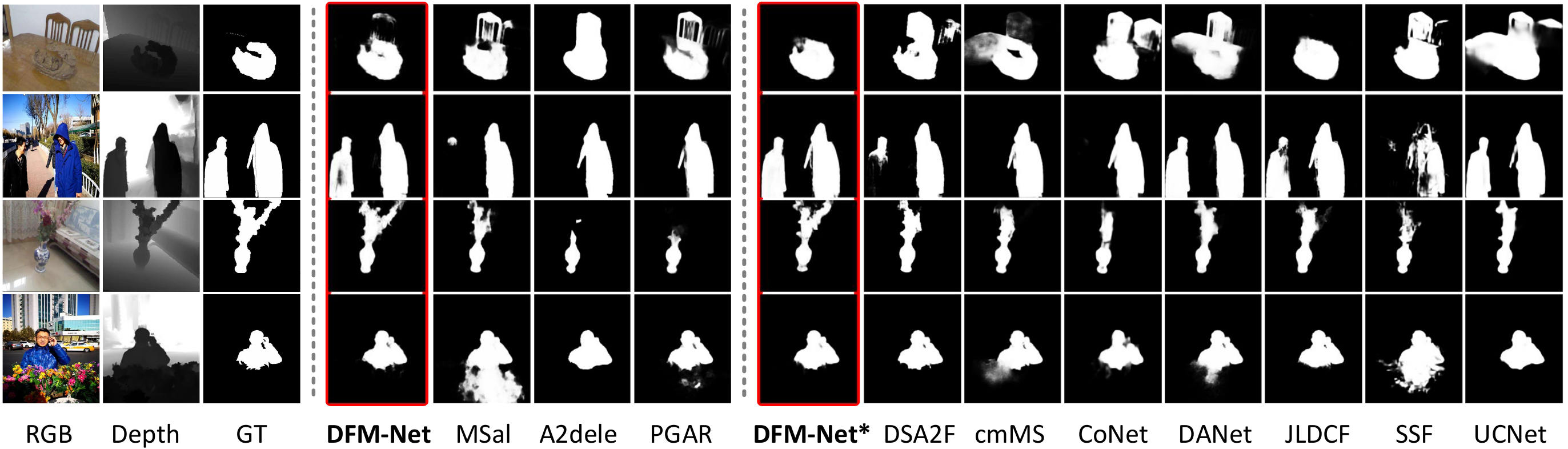,width=1.0\textwidth}}\vspace{-0cm}
\caption{Qualitative comparisons of \ourmodel~and \ourmodel* with SOTA RGB-D SOD models. Efficient and non-efficient models are shown separately.}\vspace{-0cm}
\label{fig:visual_results}
\end{figure*}

\subsection{Comparisons to SOTAs}\label{sec:comparison}
Since we cannot expect that an extremely lightweight model can always outperform existing non-efficient models which are much larger, we have also extended \ourmodel~to obtain a larger and more powerful model called \ourmodel*, by replacing MobileNet-v2 used in the RGB branch with ResNet-34~\cite{ResNet}. Then to align the features from TDB to those from the RGB branch, slight modifications are made by adjusting $c$, $t$, and $n$, finally yielding TDB* of $\sim$2.7Mb model size.

Following classic RGB-D SOD benchmark setting~\cite{JLDCF,UCNet,D3Net}, we use 1,500 samples from NJU2K and 700 samples from NLPR for training, and test on NJU2K, NLPR, STERE, RGBD135, LFSD, SIP as well as the newly released dataset RedWeb-S. Results of \ourmodel~and \ourmodel*, compared to 15 SOTA models including
DRMA \cite{DRMA}, D3Net \cite{D3Net}, JL-DCF \cite{JLDCF}, UCNet \cite{UCNet}, SSF \cite{SSF}, S2MA \cite{S2MA}, CoNet \cite{CoNet}, cmMS \cite{cmMS}, DANet \cite{DANet}, ATSA \cite{ATSA}, DCF \cite{DCF}, DSA2F \cite{DSA2F}, A2dele \cite{A2dele}, PGAR \cite{PGAR}, and MSal \cite{wu2021mobilesal} can be found in Tab. \ref{tab:benchmark}, \fkr{where the saliency
maps of other models are from their released results if available, otherwise they are computed by their released code. }
\zwb{Note that the parameter file of DSA2F \cite{DSA2F} and its results on RedWeb-S are not available, so the associated results are denoted as ``N/A''.}

\begin{table}
	\caption{Quantitative benchmark results on DUT-RGBD. Compared methods are those whose results on this dataset are publicly available, and include DMRA~\cite{DRMA}, S2MA~\cite{S2MA}, CoNet~\cite{CoNet}, DANet~\cite{DANet}, ATSA~\cite{ATSA}, DCF~\cite{DCF}, DSA2F~\cite{DSA2F}, A2dele~\cite{A2dele}, PGAR~\cite{PGAR}, and MSal~\cite{wu2021mobilesal}. Ours* and Ours refer to the proposed \ourmodel* and \ourmodel, respectively.}\vspace{-0cm}
    \tiny
	\renewcommand{\arraystretch}{1.2}
	\renewcommand{\tabcolsep}{0.46mm}
	\begin{tabular}{r|ccccccc|c||ccc|c}
		\hline\toprule
	\multirow{1}{*}{Metric} 
	
		     &\cite{DRMA}  &\cite{S2MA} &\cite{CoNet} &\cite{DANet}  &\cite{ATSA} &\cite{DCF} &\cite{DSA2F}  &Ours*&\cite{A2dele} &\cite{PGAR}&  \cite{wu2021mobilesal}&Ours \\







\midrule  $S_{\alpha}\uparrow$	&.888		&.905		&.919		&.889	&.883	&\underline{.925}	&.921	&\textbf{.924}	&.885	&\underline{.919}	&.895	&\textbf{.913}	\\
		 $F_{\beta}^{\rm m}\uparrow$	&.897		&.911				&.926		&.894	&.891	&.930	&\underline{.932}	&\textbf{.931}	&.891	&\underline{.923}	&.918	&\textbf{.917}	\\
		 $E_{\xi}^{\rm m}\uparrow$ 	&.932		&.943				&.955		&.930	&.928	&.956	&.956	&\textbf{.956}	&.928	&.948	&.949	&\textbf{.948}	\\

		 $\mathcal{M}\downarrow$	&.047		&.059			&.032		&.046	&.050	&\underline{.030}	&\underline{.030}	&\textbf{.031}	&.043	&\underline{.034}	&.044	&\textbf{.037}	\\
		\bottomrule
		\hline
	\end{tabular}
	\vspace{-0.cm}
	\label{tab:benchmark_dut-rgbd}
\end{table}

As shown in Tab. \ref{tab:benchmark}, \ourmodel~surpasses existing efficient models MSal \cite{wu2021mobilesal}, A2dele \cite{A2dele}, and PGAR \cite{PGAR} on detection accuracy and model size. It runs at 50ms on CPU, which is the fastest among all the contenders, with only $\sim$8.5Mb model size (3.1$\times$ smaller than the concurrent work MSal \cite{wu2021mobilesal}).
Regarding $S_{GPU}$, \ourmodel~exceeds the others except for A2dele \cite{A2dele} whose model size is larger than ours. This is because \ourmodel~is mainly built upon the depth-wise separable convolutions, which cannot be fully accelerated on GPU in the small batch setting. A similar issue also exists for other efficient models like MobileNet-v2 and MSal. However, regarding $S_{GPU}^*$, we can see that \ourmodel~is significantly faster than all the other models including MSal. Comparing to MSal, ours is $1.4\times$ faster.

\begin{table}[t!]
	\centering
	\caption{Quantitative benchmark results on the recent COME. Compared methods are those whose results on this dataset are publicly available, and include JL-DCF~\cite{JLDCF}, UCNet\cite{UCNet}, SSF\cite{SSF}, S2MA~\cite{S2MA}, CoNet~\cite{CoNet}, DANet~\cite{DANet}, ATSA~\cite{ATSA},  DCF~\cite{DCF}, A2dele~\cite{A2dele}, and PGAR~\cite{PGAR}. Ours* and Ours refer to the proposed \ourmodel* and \ourmodel, respectively.}\vspace{-0cm}
    \tiny
	\renewcommand{\arraystretch}{1.2}
	\renewcommand{\tabcolsep}{0.4mm}
	\begin{tabular}{lr|cccccccc|c||cc|c}
		\hline\toprule
		&\multirow{1}{*}{Metric}\centering       
	
	  &\cite{JLDCF} &\cite{UCNet}  &\cite{SSF} &\cite{S2MA} &\cite{CoNet}	&\cite{DANet}  &\cite{ATSA} &\cite{DCF}  &Ours*&\cite{A2dele} &\cite{PGAR}&Ours \\








\midrule	\multirow{4}{*}{\begin{sideways}\textit{COME-E}\end{sideways}}	&$S_{\alpha}\uparrow$			&\underline{.895}	&.891	&.852	&.876	&.837		&.884	&.807	&.892		&\textbf{.894}	&.831	&.887		&\textbf{.887}	\\
		& $F_{\beta}^{\rm m}\uparrow$			&.891	&.890	&.847	&.865	&.838		&.877	&.812	&.892		&\textbf{.895}	&.838	&.888		&\textbf{.888}	\\
		& $E_{\xi}^{\rm m}\uparrow$ 			&\underline{.932}	&\underline{.930}	&.900	&.912	&.878		&.921	&.852	&\underline{.933}		&\textbf{.926}	&.889	&.925		&\textbf{.925}	\\
		& $\mathcal{M}\downarrow$			&\underline{.041}	&\underline{.042}	&.060	&.059	&.073		&.048	&.085	&\underline{.038}		&\textbf{.045}	&.060	&.050		&\textbf{.050}	\\
\midrule	\multirow{4}{*}{\begin{sideways}\textit{COME-H}\end{sideways}}	&$S_{\alpha}\uparrow$			&\underline{.850}	&\underline{.844}	&.802	&.832	&.794		&.833	&.768	&\underline{.842}		&\textbf{.836}	&.785	&\underline{.834}		&\textbf{.830}	\\
		& $F_{\beta}^{\rm m}\uparrow$			&\underline{.848}	&\underline{.844}	&.801	&.822	&.802		&.832	&.782	&\underline{.847}		&\textbf{.838}	&.798	&\underline{.836}		&\textbf{.832}	\\
		& $E_{\xi}^{\rm m}\uparrow$ 			&\underline{.888}	&\underline{.885}	&.855	&.868	&.834		&.876	&.811	&\underline{.886}		&\textbf{.874}	&.845	&\underline{.873}		&\textbf{.872}	\\
		& $\mathcal{M}\downarrow$			&\underline{.071}	&\underline{.072}	&.094	&.091	&.105		&.080	&.116	&\underline{.070}		&\textbf{.081}	&.092	&\underline{.084}	&\textbf{.085}	\\

		\bottomrule
		\hline
	\end{tabular}
\label{tab:benchmark_come}
	\vspace{-0.cm}
\end{table}

On the other hand, one can see that \ourmodel* achieves SOTA performance when compared to existing non-efficient models, with absolutely the fastest $T_{CPU}$, $S_{GPU}$, $S_{GPU}^*$, and the smallest model size.  As early mentioned, Fig. \ref{fig:benchmark} visualizes all models' average performance across six out of the seven datasets, demonstrating the high efficiency and effectiveness of the proposed models. 

\begin{figure*}[!htb]
  \centering
 \centerline{\epsfig{figure=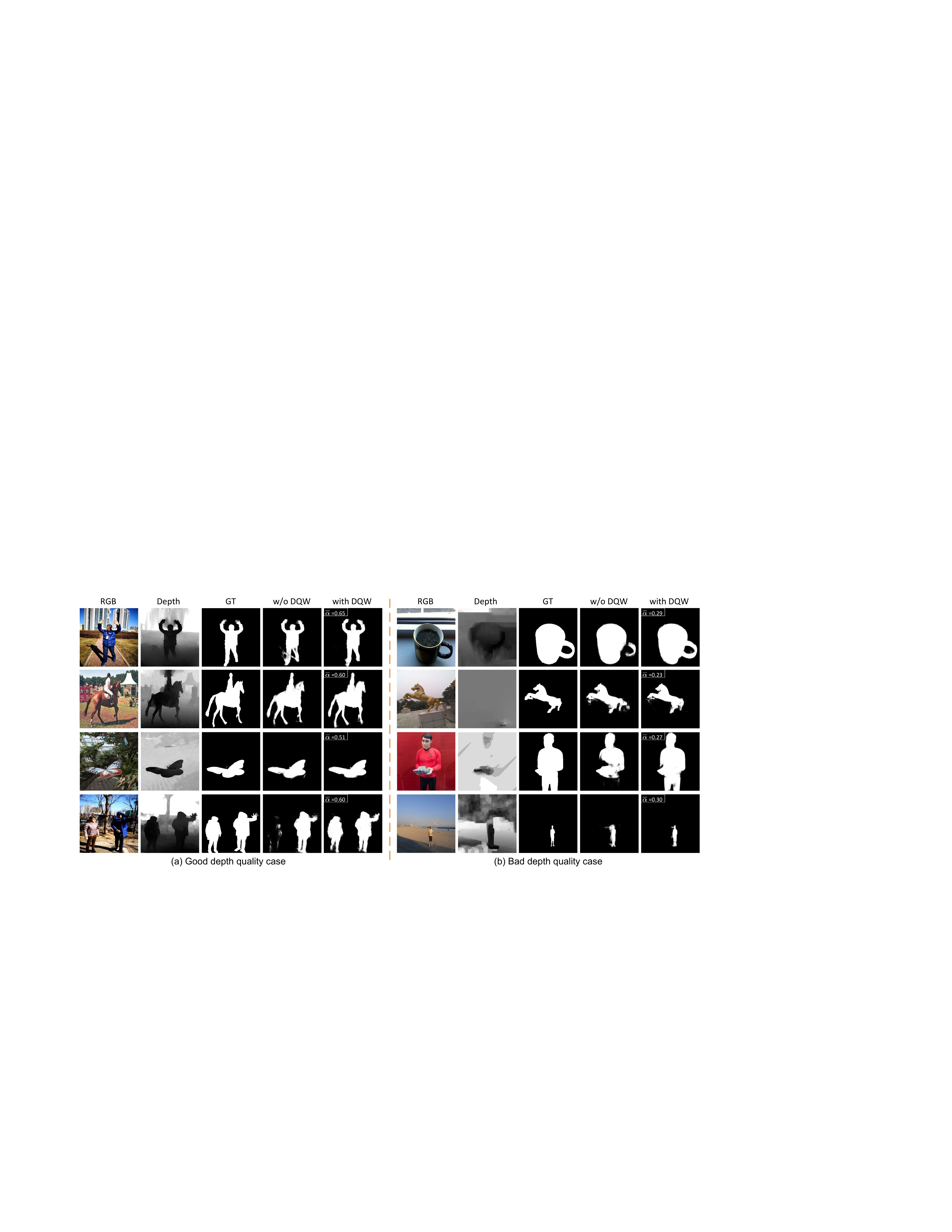,width=0.95\textwidth}}\vspace{-0cm}
\caption{Visualized examples of setting \#3 (w/o DQW) and \#4 (with DQW) for good (a) and bad (b) depth quality cases. $\overline{\alpha}$ denotes the average of $\alpha_1\sim \alpha_5$, which can be deemed as an indicator for depth quality.}
\label{fig:qualitative_dqw}
\end{figure*} 

To intuitively compare with SOTA methods, visual comparisons are further shown in Fig.~\ref{fig:visual_results}, where our results are closer to GT. Note in Fig.~\ref{fig:visual_results}, efficient and non-efficient models are compared separately.

\begin{table*}[htbp]
	\centering
 	\tiny
	\renewcommand{\arraystretch}{1.1}
	\renewcommand{\tabcolsep}{0.4mm}
	\caption{Ablation analyses for DQFM, where the effectiveness of DQW and DHA is validated. The best results are highlighted in \textbf{bold}.}\vspace{-0cm}
	\label{tab:ablation_DQFM}
	\begin{tabular}{c|c|c|c|cccc|cccc|cccc|cccc|cccc|cccc}
		\hline\toprule
		\multirow{2}{*}{\#} 
		&\multirow{2}{*}{DQW}
		&\multirow{2}{*}{DHA}
		&Size
		&\multicolumn{4}{c|}{\textbf{SIP~\cite{D3Net}}}  &\multicolumn{4}{c|}{\textbf{NLPR~\cite{NLPR}}}
		&\multicolumn{4}{c|}{\textbf{NJU2K~\cite{NJU2K}}}
		&\multicolumn{4}{c|}{\textbf{RGBD135~\cite{RGBD135}}}
		&\multicolumn{4}{c|}{\textbf{LFSD~\cite{LFSD}}}
		&\multicolumn{4}{c}{\textbf{STERE~\cite{STERE}}}\\ &&&(Mb)
        &$S_{\alpha}$
	    &$F_{\beta}^{\rm m}$
	    & $E_{\xi}^{\rm m}$ 
		&$\mathcal{M}$  
		&$S_{\alpha}$
	    &$F_{\beta}^{\rm m}$
	    & $E_{\xi}^{\rm m}$ 
		&$\mathcal{M}$  
		&$S_{\alpha}$
	    &$F_{\beta}^{\rm m}$
	    & $E_{\xi}^{\rm m}$ 
		&$\mathcal{M}$  
		&$S_{\alpha}$
	    &$F_{\beta}^{\rm m}$
	    & $E_{\xi}^{\rm m}$ 
		&$\mathcal{M}$  
		&$S_{\alpha}$
	    &$F_{\beta}^{\rm m}$
	    & $E_{\xi}^{\rm m}$ 
		&$\mathcal{M}$  
		&$S_{\alpha}$
	    &$F_{\beta}^{\rm m}$
	    & $E_{\xi}^{\rm m}$ 
		&$\mathcal{M}$\\
		\hline
1	&	&&8.409	&.849	&.842	&.897	&.070	&.907	&.884	&.946	&.032	&.890	&.887	&.931	&.052	&.929	&.915	&.968	&.025	&.847	&.841	&.883	&.084	&.880	&.872	&.931	&.054	\\

2	&\checkmark 	& &	8.416	&.878	&.884	&.922	&.054	&.918	&.902	&.957	&.027	&.898	&.898	&.938	&.047	&\textbf{{.932}}	&.922	&.971	&.022	&.857	&.856	&.896	&.078	&.894	&.886	&.939	&.047	\\ 

3	&	&\checkmark 	&8.451		&.861	&.861	&.908	&.063	&.918	&.901	&.953	&.028	&.899	&.896	&.940	&.046	&.921	&.911	&.966	&.024	&.857	&.855	&.895	&.076	&.897	&.891	&\textbf{{.942}}	&.045	\\

\rowcolor{mygray}

4	&\checkmark 	&\checkmark 		&8.459	&\textbf{{.883}}	&\textbf{{.887}}	&\textbf{{.926}}	&\textbf{{.051}}	&\textbf{{.923}}	&\textbf{{.908}}	&\textbf{{.957}}	&\textbf{{.026}}	&\textbf{{.906}}  &\textbf{{.910}}	&\textbf{{.947}}	&\textbf{{.042}}	&.931	&\textbf{{.922}}	&\textbf{{.972}}	&\textbf{{.021}}	&\textbf{{.865}}	&\textbf{{.864}}	&\textbf{{.903}}	&\textbf{{.072}}	&\textbf{{.898}}	&\textbf{{.893}}	&.941	&\textbf{{.045}}	\\



         







		\bottomrule
		\hline
	\end{tabular}
	\vspace{-0.2cm}
\end{table*}

\subsection{Additional Evaluation on DUT-RGBD and COME}
For more comprehensive evaluation, we test our methods on two extra datasets, namely DUT-RGBD \cite{DRMA} and COME \cite{zhang2021rgb}. To evaluate on DUT-RGBD, we follow the benchmark setting as previous works~\cite{CoNet,PGAR,DANet,ATSA,A2dele}, where 800 samples from DUT-RGBD are also added for training besides the training data from NJU2K and NLPR. Fairly, all the compared models are trained under this setting and publicly provided by the authors. Testing is conducted on 400 samples remained in DUT-RGBD, and the results are shown in Tab.~\ref{tab:benchmark_dut-rgbd}. It can be seen that without bells and whistles, \ourmodel~and \ourmodel* achieve better balance between accuracy and efficiency, and are 4.7/7.2$\times$ smaller than the most accurate efficient/non-efficient ones (\cite{DSA2F}/\cite{PGAR}) with little accuracy degradation.

Recently, Zhang\etal{} \cite{zhang2021rgb} introduces a new RGB-D SOD benchmark called COME, which officially splits out 8,025 samples for training and the rest samples are divided into COME-E and COME-H for normal and hard cases. COME-E and COME-H include 4,600 and 3,000 samples, respectively. We re-train our methods on its training set, and the compared models are the re-trained off-the-shelf versions provided by Zhang\etal{} in their benchmark \cite{zhang2021rgb}. As shown in Tab.~\ref{tab:benchmark_come}, the  competitive performance of \ourmodel~and \ourmodel*~again illustrates their adaptability onto this brand new benchmark dataset.

\subsection{Ablation Studies}\label{sec:ablation}
We conduct thorough ablation studies on six classic datasets including NJU2K, NLPR, STERE, RGBD135, LFSD, and SIP, by removing or replacing components from the full implementation of \ourmodel.

\begin{figure*}
  \centering
 \centerline{\epsfig{figure=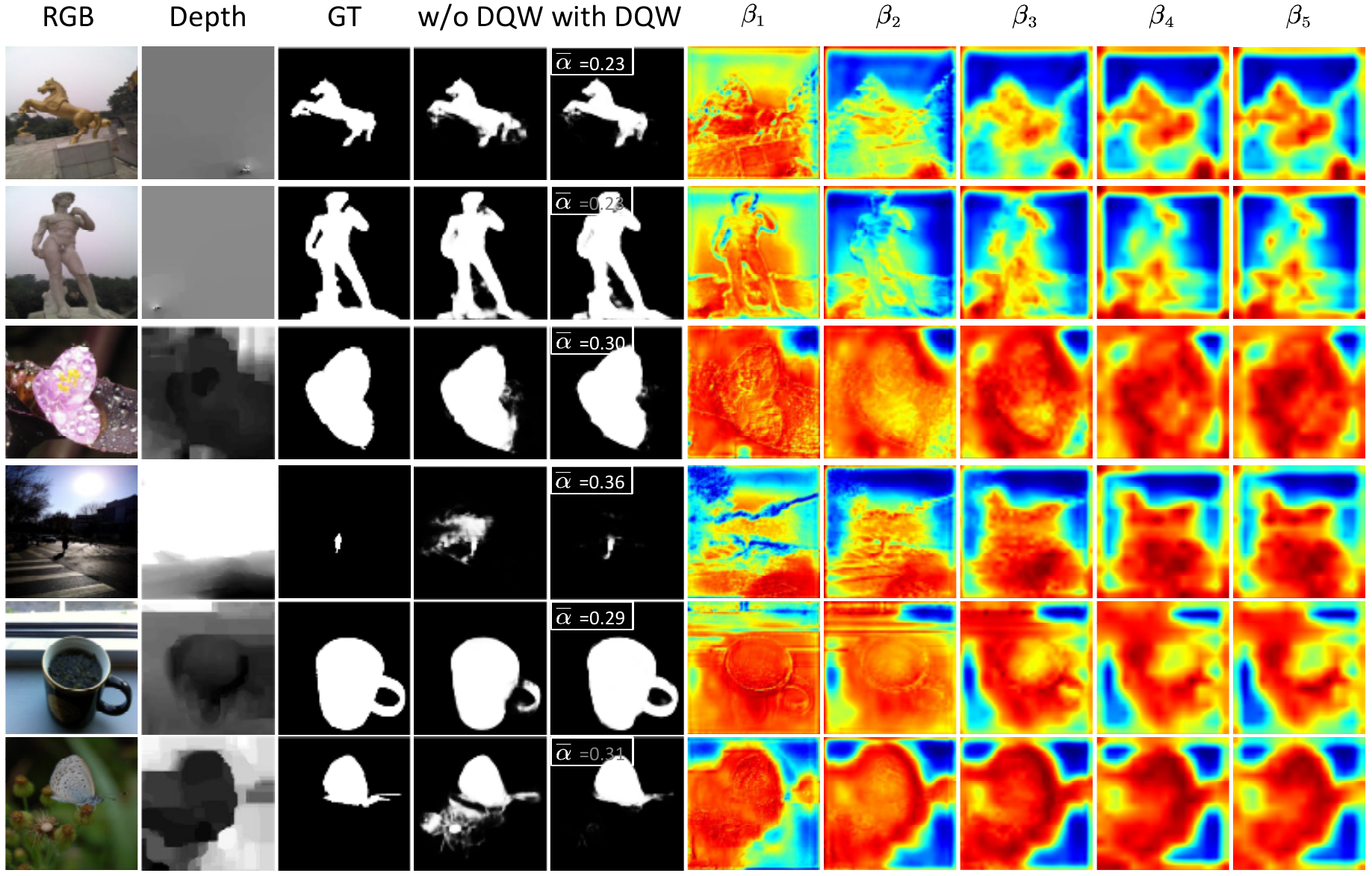,width=0.95\textwidth}}\vspace{-0cm}
\caption{Saliency maps with and without DQW, and the holistic attention maps $\beta_1 \sim \beta_5$ achieved by DHA. $\overline{\alpha}$ achieved by DQW is also shown. \fkr{Note the somewhat inaccurate holistic attention maps generated by DHA, which are derived from the low-quality depth maps.}}
\label{fig:synergy}
\vspace{0cm}
\end{figure*}

\subsubsection{Effectiveness of DQFM} DQFM consists of two key components, namely DQW and DHA. Tab.~\ref{tab:ablation_DQFM} shows different configurations by ablating DQW/DHA. In detail, \#1 denotes a baseline model which has removed both DQW and DHA from \ourmodel. Configuration \#2 and \#3 mean having either one component, while \#4 means the full model of \ourmodel. 
Basically, from Tab.~\ref{tab:ablation_DQFM} one can see that incorporating either DQW and DHA into the baseline model \#1 leads to consistent improvement on almost all datasets. Besides, comparing \#2/\#3 to \#4, we see that employing both DQW and DHA can further enhance the results, demonstrating the synergy effect between DQW and DHA. The underlying reason could be that, although DHA is able to enhance potential target regions in the depth, it is unavoidable to make some mistakes (\emph{e.g.}, highlight wrong regions) especially in low depth quality cases. Luckily, DQW somewhat relieves such a side-effect because in the meantime it assigns low global weights to depth features. In all, these two components can work cooperatively to improve the robustness of the network, as we mentioned in Sec.~\ref{sec:DQFM}. Last but not least, from the model sizes of different configurations shown in Tab.~\ref{tab:ablation_DQFM}, the introduction of DQW and DHA brings merely $\sim$0.6\% extra parameters compared to the baseline \#1. This rightly meets our goal and implies that they could potentially become universal components for lightweight models in the future.

\begin{table*}[htbp]
	\centering
	\caption{Validation of $V_{BA}$ in DQW, including the proposed operation (Pro.), original Dice-like computation (Dice.), straightforward addition (Add.), and straightforward multiplication (Mul.). Note we have added the last ``Average'' group, which shows the average scores over those six benchmark datasets, in order for comparison convenience. The best results are highlighted in \textbf{bold}.}\vspace{-0cm}
	\label{tab:ablation-bastrategy}
	
    \tiny
	\renewcommand{\arraystretch}{0.9}
	\renewcommand{\tabcolsep}{0.2mm}
	\begin{tabular}{c|c|cccc|cccc|cccc|cccc|cccc|cccc|cccc}
		\hline\toprule
		\multirow{2}{*}{\#} 

        &\multirow{2}{*}{$V_{BA}$}

		&\multicolumn{4}{c|}{\textbf{SIP~\cite{D3Net}}}  &\multicolumn{4}{c|}{\textbf{NLPR~\cite{NLPR}}}
		&\multicolumn{4}{c|}{\textbf{NJU2K~\cite{NJU2K}}}
		&\multicolumn{4}{c|}{\textbf{RGBD135~\cite{RGBD135}}}
		&\multicolumn{4}{c|}{\textbf{LFSD~\cite{LFSD}}}
		&\multicolumn{4}{c|}{\textbf{STERE~\cite{STERE}}}
		&\multicolumn{4}{c}{\textbf{Average}}\\ &&
        $S_{\alpha}$
	    &$F_{\beta}^{\rm m}$
	    & $E_{\xi}^{\rm m}$ 
		&$\mathcal{M}$  
		&$S_{\alpha}$
	    &$F_{\beta}^{\rm m}$
	    & $E_{\xi}^{\rm m}$ 
		&$\mathcal{M}$  
		&$S_{\alpha}$
	    &$F_{\beta}^{\rm m}$
	    & $E_{\xi}^{\rm m}$ 
		&$\mathcal{M}$  
		&$S_{\alpha}$
	    &$F_{\beta}^{\rm m}$
	    & $E_{\xi}^{\rm m}$ 
		&$\mathcal{M}$  
		&$S_{\alpha}$
	    &$F_{\beta}^{\rm m}$
	    & $E_{\xi}^{\rm m}$ 
		&$\mathcal{M}$  
		&$S_{\alpha}$
	    &$F_{\beta}^{\rm m}$
	    & $E_{\xi}^{\rm m}$ 
		&$\mathcal{M}$
		&$S_{\alpha}$
	    &$F_{\beta}^{\rm m}$
	    & $E_{\xi}^{\rm m}$ 
		&$\mathcal{M}$
	   
\\\midrule

		\rowcolor{mygray}
		4 &Pro.  &\textbf{{.883}}	&\textbf{{.887}}	&{.926}	&\textbf{{.051}}	&\textbf{{.923}}	&{{.908}}	&{.957}	&\textbf{.026}	&\textbf{{.906}}  &\textbf{{.910}}	&\textbf{.947}	&\textbf{.042}	&.931	&.922	&.972	&\textbf{.021}	&.865	&\textbf{{.864}}	&\textbf{{.903}}	&{{.072}}	&{{.898}}	&{{.893}}	&.941	&{{.045}} &\textbf{.901}	&\textbf{.897}	&\textbf{.941}	&\textbf{.043}
\\
5	&Dice.	&.877	&.881	&.922	&.053	&.921	&.907	&.957	&.026	&.904	&.904	&.942	&.044	&.926	&.917	&.969	&.023	&\textbf{.866}	&.861	&.901	&\textbf{.070}	&.897	&.891	&.942	&.045  &.898	&.893	&.939	&.044
\\
6	&Add. &.870	&.873	&.916	&.057	&.921	&\textbf{.909}	&\textbf{.959}	&.026	&.904	&.904	&.944	&.044	&.922	&.911	&.968	&.025	&.859	&.856	&.895	&.074	&.900	&.893	&.942	&.044  &.896	&.891	&.937	&.045

	\\
7	&Mul. &.881	&.885	&\textbf{.927}	&.051	&.920	&.907	&.956	&.027	&.902	&.900	&.943	&.044	&\textbf{.933}	&\textbf{.924}	&\textbf{.973}	&.021	&.862	&.859	&.898	&.072	&\textbf{.902}	&\textbf{.896}	&\textbf{.946}	&\textbf{.042} &.900	&.895	&.940	&.043

	\\

		\bottomrule
		\hline
	\end{tabular}
	\vspace{-0.cm}
\end{table*}

\begin{figure*}[htbp]
  \centering
 \centerline{\epsfig{figure=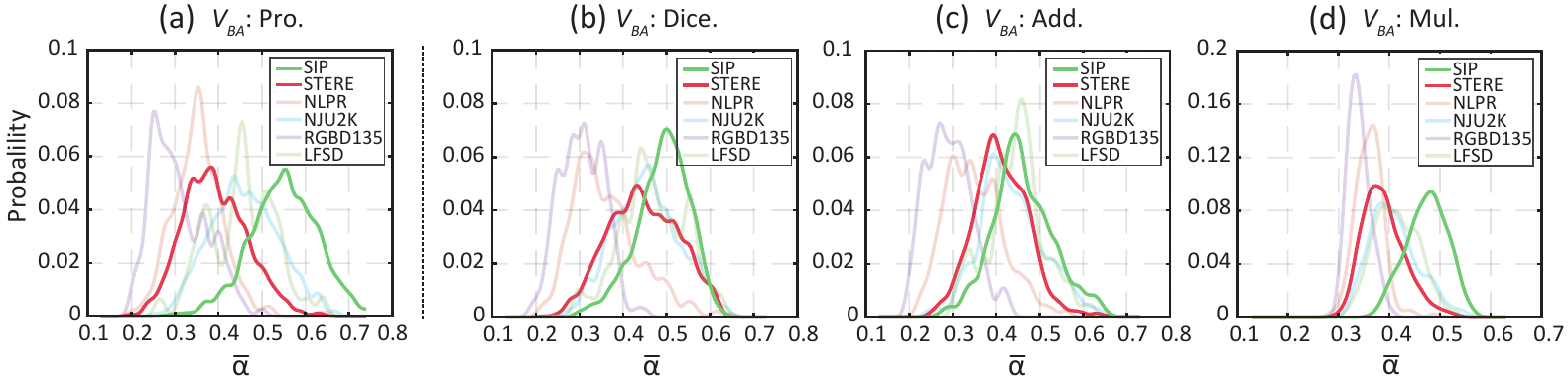,width=1\textwidth}}
\caption{Distribution of the inferred $\overline{\alpha}$. (a) denotes the distribution corresponding to the proposed operation for $V_{BA}$. (b)-(d) denote different variants applied for $V_{BA}$, namely the original Dice coefficient-like computation, straightforward addition, and straightforward multiplication. \zwb{Note the colors of the distributions on STERE and SIP are emphasized for better comparison.}}
\label{fig:alpha_distribution}
\end{figure*}

In Fig.~\ref{fig:qualitative_dqw}, we show visual examples of setting \#3 (namely without DQW) and \#4. We also visualize the magnitudes of $\alpha$, namely the mean value $\overline{\alpha}$ of $\alpha_1 \sim \alpha_5$. From Fig.~\ref{fig:qualitative_dqw} (a) and (b), we can see that incorporating DQW does help improve detection accuracy, and practically, DQW is able to function as expected, namely rendering the good quality cases with higher weights (Fig.~\ref{fig:qualitative_dqw} (a)), and vice versa (Fig.~\ref{fig:qualitative_dqw} (b)). In the first (row 1) good quality case of (a), it is difficult to distinguish between the shadow and the man's legs in the RGB view, but this can be done easily in the depth view. Incorporating DQW to give more emphasis to depth features, therefore, can help better separate the entire human body from the shadow. In the first (also row 1) bad quality case of (b), although in the depth view the cup handle is much blurry, the impact of misleading depth has been alleviated, and the whole object still can be detected out accurately.

\fkr{
We also validate the synergy effect of combining DQW and DHA. Since DHA cannot always highlight correct regions, what if DHA gives wrong response? We display in Fig. \ref{fig:synergy} some cases where DHA generates misleading maps and show such effects can be alleviated by DQW (with lower $\overline{\alpha}$ achieved). This proves that DHA and DQW could cooperate with each other and enhance detection robustness. That is, incorporating DQW into DHA (comparing between ``with DQW'' and ``w/o DQW'' columns in Fig. \ref{fig:synergy}) consistently improves the detection accuracy. 
}

\subsubsection{Delving into DQW} 
The above only give intuitive examples, but \emph{does DQW really function as we expected?} To statistically validate the role of DQW, we plot the distribution curves of the inferred $\overline{\alpha}$ on different datasets. As shown in Fig. \ref{fig:alpha_distribution} (a), we can see that DQW can distinguish between SIP and STERE well by the predicted $\overline{\alpha}$, as expected by our BA motivation in Fig. \ref{fig:distribution}. In addition to this, we also validate other possible variants of Eq.~\ref{eq:4} in DQW, including straightforward addition, multiplication, as well as the original Dice coefficient-like computation (with square in the denominator). These variants are denoted as \#6, \#7, \#5, respectively. Tab. \ref{tab:ablation-bastrategy} compares the performance of these variants and their $\overline{\alpha}$ are exhibited in Fig. \ref{fig:alpha_distribution} (b)-(d).

From Tab.~\ref{tab:ablation-bastrategy}, one can see according to the average evaluation scores that the proposed operation performs better than the other variants, including the original Dice coefficient-like computation. We find that this phenomenon should be attributed to the fact that the proposed operation results in a more reasonable distribution of $\overline{\alpha}$, and can  
better indicate the low-level alignment between RGB and depth.
It is also interesting to note that, \#7 (``Mul.'') practically performs better than the remaining two variants, and is very close to the performance of \#4. The underlying reasons may be two-fold: 1) Though without normalization by the denominator, using merely feature multiplication still can somewhat excavate and capture the common low-level edge features. 2) The obtained distribution of $\overline{\alpha}$ from \#7, namely Fig. \ref{fig:alpha_distribution} (d), also seems to be more reasonable and can better tell STERE and SIP apart.

\begin{table*}[!htb]
	\centering
	\caption{Ablation analyses for times of recalibration process $\textbf{F}_{rec}$ in DHA. Zero means no recalibration is conducted.}\vspace{-0cm}
	\label{tab:ablation2}
    \tiny
	\renewcommand{\arraystretch}{0.9}
	\renewcommand{\tabcolsep}{0.68mm}
	\begin{tabular}{c|c|cccc|cccc|cccc|cccc|cccc|cccc}
		\hline\toprule
		\multirow{2}{*}{\#} 
		&\multirow{2}{*}{$\textbf{F}_{rec}$}
		&\multicolumn{4}{c|}{\textbf{SIP~\cite{D3Net}}}  &\multicolumn{4}{c|}{\textbf{NLPR~\cite{NLPR}}}
		&\multicolumn{4}{c|}{\textbf{NJU2K~\cite{NJU2K}}}
		&\multicolumn{4}{c|}{\textbf{RGBD135~\cite{RGBD135}}}
		&\multicolumn{4}{c|}{\textbf{LFSD~\cite{LFSD}}}
		&\multicolumn{4}{c}{\textbf{STERE~\cite{STERE}}}\\ &
        &$S_{\alpha}$
	    &$F_{\beta}^{\rm m}$
	    & $E_{\xi}^{\rm m}$ 
		&$\mathcal{M}$  
		&$S_{\alpha}$
	    &$F_{\beta}^{\rm m}$
	    & $E_{\xi}^{\rm m}$ 
		&$\mathcal{M}$  
		&$S_{\alpha}$
	    &$F_{\beta}^{\rm m}$
	    & $E_{\xi}^{\rm m}$ 
		&$\mathcal{M}$  
		&$S_{\alpha}$
	    &$F_{\beta}^{\rm m}$
	    & $E_{\xi}^{\rm m}$ 
		&$\mathcal{M}$  
		&$S_{\alpha}$
	    &$F_{\beta}^{\rm m}$
	    & $E_{\xi}^{\rm m}$ 
		&$\mathcal{M}$  
		&$S_{\alpha}$
	    &$F_{\beta}^{\rm m}$
	    & $E_{\xi}^{\rm m}$ 
		&$\mathcal{M}$
	   
\\\midrule
8 &0&.877	&.881	&.919	&.054	&.920	&.904	&.953	&.027	&.904	&.904	&.943	&.043	&\textbf{.934}	&\textbf{.924}	&\textbf{.973}	&.021	&.855	&.855	&.894	&.076	&.897	&.890	&.940	&.045\\
9 &1&.872	&.873	&.918	&.056	&.920	&.904	&.957	&.027	&.903	&.902	&.943	&.043	&.929	&.922	&.973	&.023	&.864	&\textbf{.865}	&.902	&.072	&.900	&.893	&.942	&.044

\\
		\rowcolor{mygray}
		4 &2 &\textbf{{.883}}	&\textbf{{.887}}	&\textbf{{.926}}	&\textbf{{.051}}	&\textbf{{.923}}	&\textbf{{.908}}	&\textbf{.957}	&\textbf{{.026}}	&\textbf{{.906}}  &\textbf{{.910}}	&\textbf{{.947}}	&\textbf{{.042}}	&.931	&.922	&.972	&\textbf{.021}	&\textbf{{.865}}	&.864	&\textbf{{.903}}	&\textbf{{.072}}	&.898	&.893	&.941	&.045
\\
10 &3 &.864 &.868  &.916 &.059&.921  &.907  &.957 &.026 &.898  &.897  &.941  &.042	 	   &.917  &.905  &.963  &.025   &.858  &.858  &.896  &.077  &\textbf{.905}  &\textbf{.897}  &\textbf{.946}  &\textbf{.042}
\\
		\bottomrule
		\hline
	\end{tabular}
	\vspace{-0.cm}
	\label{tab:recalibration}
\end{table*}

\begin{figure*}[!htb]
  \centering
 \centerline{\epsfig{figure=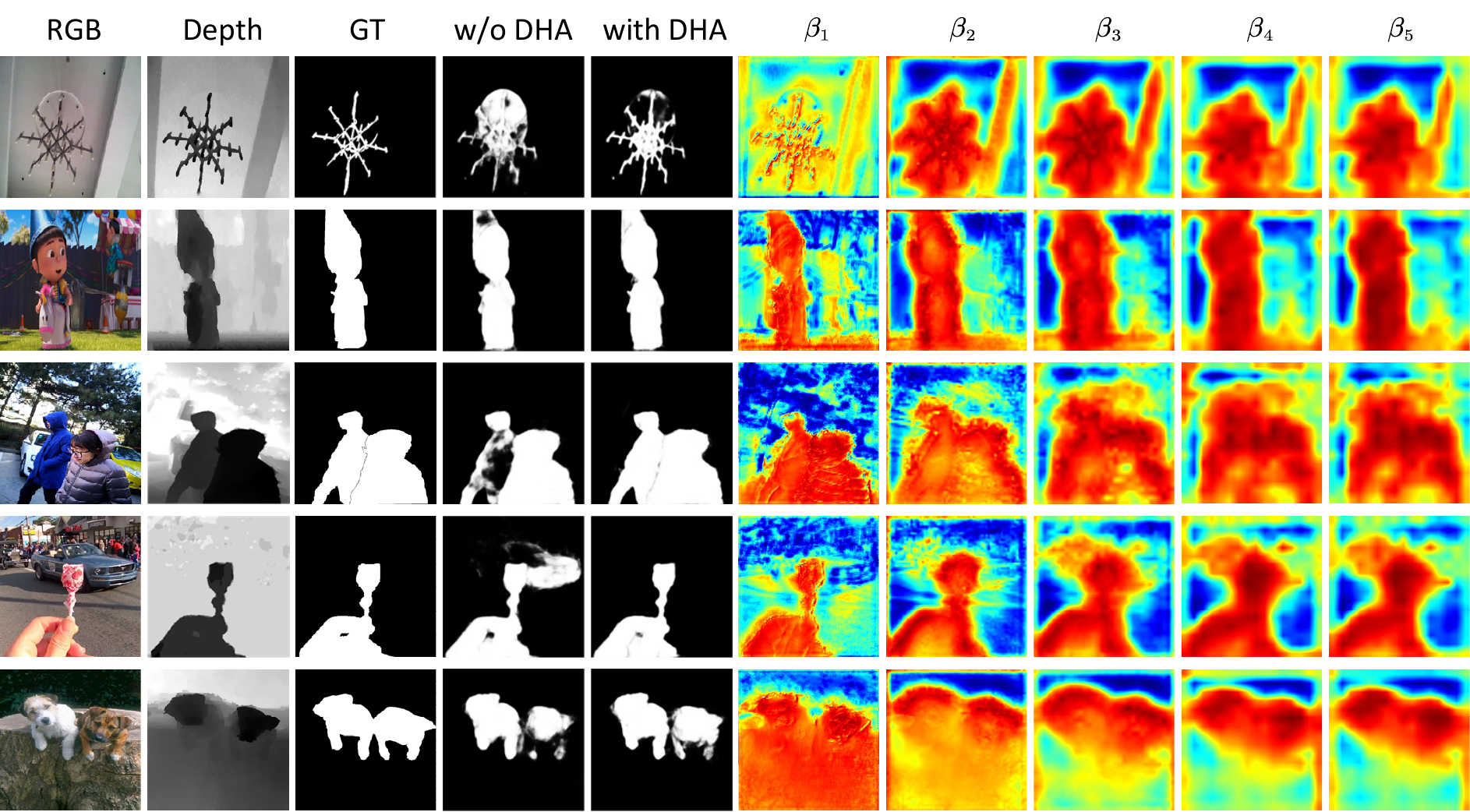,width=0.95\textwidth}}\vspace{-0cm}
\caption{Saliency maps with and without DHA, and the holistic attention maps $\beta_1 \sim \beta_5$ achieved by DHA.}
\label{fig:dhamore}
\vspace{0cm}
\end{figure*}

However, we notice that all the variants, including the proposed operation, apparently fail on RGBD135, because the obtained $\overline{\alpha}$ distribution for RGBD135 is on the left of STERE. This means that the individual model thinks RGBD135 presents worse depth quality than STERE. However, according to the observation of human eyes and also previous works \cite{SMAC,D3Net,JLDCF}, the general depth quality of RGBD135 is not that bad and should be better than that of STERE. This
failure case of DQW on RGBD135 may also explain why DQW has the least improvement for this dataset when comparing \#2 with \#1 in Tab.~\ref{tab:ablation_DQFM}. We attribute this failure case to the unsupervised learning mechanism of DQW, which means that without explicit guidance, the rules implicitly learned from data may not be that accurate and cannot generalize well to some unseen scenarios. Interestingly, we observe improved performance on RGBD135 for \ourmodel~if we manually modify the obtained $\alpha$ further to larger values during inference.    

Finally, we notice that our inferred depth quality results of datasets, which are indicated by $\overline{\alpha}$, are partially consistent with the analyses in a recent work~\cite{LSMA}. In~\cite{LSMA}, the authors point out that the descending order for depth map quality of existing datasets is: SIP, NLPR, RGBD135, NJU2K, LFSD, and STERE. From Fig. \ref{fig:alpha_distribution} (a), we can derive that the ranking statistic from our \ourmodel~is: SIP, NJU2K, LFSD, STERE, NLPR, and RGBD135, which coincides with the former if NLPR and RGBD135 are ignored. We believe such a coincidence, from the other aspect, validates the feasibility and effectiveness of our proposed model.          


\begin{table*}[htbp]
	\centering
	\caption{DQFM gating strategy: using identical (only one) $\alpha_i$ and $\beta_i$ \emph{vs.} using multiple (five different) $\alpha_i$ and $\beta_i$. }\vspace{-0cm}
	\label{tab:ablation3}
    \tiny
	\renewcommand{\arraystretch}{0.9}
	\renewcommand{\tabcolsep}{0.55mm}
	\begin{tabular}{c|c|cccc|cccc|cccc|cccc|cccc|cccc}
		\hline\toprule
		\multirow{2}{*}{\#} 
		&\multirow{2}{*}{Strategy}
		&\multicolumn{4}{c|}{\textbf{SIP~\cite{D3Net}}}  &\multicolumn{4}{c|}{\textbf{NLPR~\cite{NLPR}}}
		&\multicolumn{4}{c|}{\textbf{NJU2K~\cite{NJU2K}}}
		&\multicolumn{4}{c|}{\textbf{RGBD135~\cite{RGBD135}}}
		&\multicolumn{4}{c|}{\textbf{LFSD~\cite{LFSD}}}
		&\multicolumn{4}{c}{\textbf{STERE~\cite{STERE}}}\\ &
        &$S_{\alpha}$
	    &$F_{\beta}^{\rm m}$
	    & $E_{\xi}^{\rm m}$ 
		&$\mathcal{M}$  
		&$S_{\alpha}$
	    &$F_{\beta}^{\rm m}$
	    & $E_{\xi}^{\rm m}$ 
		&$\mathcal{M}$  
		&$S_{\alpha}$
	    &$F_{\beta}^{\rm m}$
	    & $E_{\xi}^{\rm m}$ 
		&$\mathcal{M}$  
		&$S_{\alpha}$
	    &$F_{\beta}^{\rm m}$
	    & $E_{\xi}^{\rm m}$ 
		&$\mathcal{M}$  
		&$S_{\alpha}$
	    &$F_{\beta}^{\rm m}$
	    & $E_{\xi}^{\rm m}$ 
		&$\mathcal{M}$  
		&$S_{\alpha}$
	    &$F_{\beta}^{\rm m}$
	    & $E_{\xi}^{\rm m}$ 
		&$\mathcal{M}$
	   
\\\midrule
		11 &Identical &.880	&.884	&.924	&.053	&.922	&.907	&\textbf{{.958}}	&.026	&.901	&.901	&.944	&.044	&.924	&.911	&.961	&.024	&.859	&.857	&.899	&.073	&.898	&.892	&\textbf{{.942}}	&.045
		
\\
		\rowcolor{mygray}
		4 &Multiple &\textbf{{.883}}	&\textbf{{.887}}	&\textbf{{.926}}	&\textbf{{.051}}	&\textbf{{.923}}	&\textbf{{.908}}	&.957	&\textbf{{.026}}	&\textbf{{.906}}  &\textbf{{.910}}	&\textbf{{.947}}	&\textbf{{.042}}	&\textbf{{.931}}	&\textbf{{.922}}	&\textbf{{.972}}	&\textbf{{.021}}	&\textbf{{.865}}	&\textbf{{.864}}	&\textbf{{.903}}	&\textbf{{.072}}	&\textbf{{.898}}	&\textbf{{.893}}	&.941	&\textbf{{.045}}
\\

		\bottomrule
		\hline
	\end{tabular}
	\vspace{-0.cm}
\end{table*}

\begin{table*}[htbp]
	\centering
	\caption{Performance of the proposed TDB (tailored depth backbone) against MobileNet-V2. Details are in Sec.~\ref{sec:ablation}.}\vspace{-0cm}
	\label{tab:ablation4}
	
    \tiny
	\renewcommand{\arraystretch}{0.9}
	\renewcommand{\tabcolsep}{0.36mm}
	\begin{tabular}{c|c|c|cccc|cccc|cccc|cccc|cccc|cccc}
		\hline\toprule
		\multirow{2}{*}{\#} 

        &Depth
		&Size

		&\multicolumn{4}{c|}{\textbf{SIP~\cite{D3Net}}}  &\multicolumn{4}{c|}{\textbf{NLPR~\cite{NLPR}}}
		&\multicolumn{4}{c|}{\textbf{NJU2K~\cite{NJU2K}}}
		&\multicolumn{4}{c|}{\textbf{RGBD135~\cite{RGBD135}}}
		&\multicolumn{4}{c|}{\textbf{LFSD~\cite{LFSD}}}
		&\multicolumn{4}{c}{\textbf{STERE~\cite{STERE}}}\\ &backbone &(Mb)
        &$S_{\alpha}$
	    &$F_{\beta}^{\rm m}$
	    & $E_{\xi}^{\rm m}$ 
		&$\mathcal{M}$  
		&$S_{\alpha}$
	    &$F_{\beta}^{\rm m}$
	    & $E_{\xi}^{\rm m}$ 
		&$\mathcal{M}$  
		&$S_{\alpha}$
	    &$F_{\beta}^{\rm m}$
	    & $E_{\xi}^{\rm m}$ 
		&$\mathcal{M}$  
		&$S_{\alpha}$
	    &$F_{\beta}^{\rm m}$
	    & $E_{\xi}^{\rm m}$ 
		&$\mathcal{M}$  
		&$S_{\alpha}$
	    &$F_{\beta}^{\rm m}$
	    & $E_{\xi}^{\rm m}$ 
		&$\mathcal{M}$  
		&$S_{\alpha}$
	    &$F_{\beta}^{\rm m}$
	    & $E_{\xi}^{\rm m}$ 
		&$\mathcal{M}$
	   
\\\midrule
		12 &MoblieNet-V2 &6.9 &.879	&.886	&.923	&.054	&.919	&.904	&.954	&.027	&.906	&.908	&\textbf{{.948}}	&.042	&.930	&.922	&.969	&.022	&.864	&.864	&.902	&.070	&.893	&.889	&\textbf{{.942}}	&.046
		
\\
		\rowcolor{mygray}
		4 &Tailored &0.9 &\textbf{{.883}}	&\textbf{{.887}}	&\textbf{{.926}}	&\textbf{{.051}}	&\textbf{{.923}}	&\textbf{{.908}}	&\textbf{{.957}}	&\textbf{{.026}}	&\textbf{{.906}}  &\textbf{{.910}}	&.947	&\textbf{{.042}}	&\textbf{{.931}}	&\textbf{{.922}}	&\textbf{{.972}}	&\textbf{{.021}}	&\textbf{{.865}}	&\textbf{{.864}}	&\textbf{{.903}}	&\textbf{{.072}}	&\textbf{{.898}}	&\textbf{{.893}}	&.941	&\textbf{{.045}}

\\

		\bottomrule
		\hline
	\end{tabular}
	\vspace{-0.cm}
\end{table*}
\subsubsection{Recalibration in DHA}\label{sec:ablation_DHA} As described in Sec.~\ref{sec:DHA}, in DHA, we utilize operation $\textbf{F}_{rec}$ to recalibrate the coarse information from high-level depth features. To validate the necessity of using $\textbf{F}_{rec}$, we experiment with different times (from 0 to 3) of using $\textbf{F}_{rec}$. These variants are denoted as \#5, \#6, \#4, and \#7 in Tab.~\ref{tab:recalibration}. Note that \#4 corresponds to the default implementation of \ourmodel. From Tab. \ref{tab:recalibration}, 
we can see that \#4 (recalibrate twice) achieves the overall best performance. The underlying reason should be that, appropriate usage of $\textbf{F}_{rec}$ can expand the coverage areas of attention maps to make conservative filtering for object edges as well as some inaccurately located objects, but too large receptive field leads to over-dilated attention regions that are less informative. This is the reason why when the times increase to 3, the performance starts to degenerate on most datasets, except on STERE whose depth quality is generally low, which easily leads to inaccurate attention location. \fkr{In Fig.~\ref{fig:dhamore} we show more visual examples by using/removing DHA modules to intuitively show its function.}

\subsubsection{DQFM Gating Strategy}
As we mentioned in Sec.~\ref{sec:DQFM}, we adopt a multi-variable strategy for $\alpha_i$ and $\beta_i$. To validate this strategy, we compare it to the single-variable strategy, namely using identical (only one) $\alpha_i$ and $\beta_i$. Tab.~\ref{tab:ablation3} shows the results, from which it can be seen that our proposed multi-variable strategy is better, because it somewhat increases the network flexibility by rendering different hierarchies with different quality-inspired weights and attention maps.

\subsubsection{Tailored Depth Backbone}The effectiveness of TDB is validated by comparing it to MobileNet-V2. We implement a configuration \#9 by switching TDB directly to MobileNet-V2, while maintaining all other settings unchanged. Evaluation results are shown in Tab. \ref{tab:ablation4}. \zwb{We can see that our tailored depth backbone is much lighter (0.9Mb vs. MobileNet-V2's 6.9Mb) but able to achieve slightly better accuracy when compared to MobileNet-V2 for \ourmodel. Although this phenomenon seems counter-intuitive, it is not unique. As a recent large benchmark of model robustness~\cite{tang2021robustart} shows, for some lightweight model families such as MobileNet-V2, MobileNet-V3, and ShuffleNet-V2, larger versions are not necessarily more robust than the smaller ones. 
This sheds light on the possibility of utilizing a lighter backbone to process depth data for efficiency purpose.} 

\begin{figure}
  \centering
 \centerline{\epsfig{figure=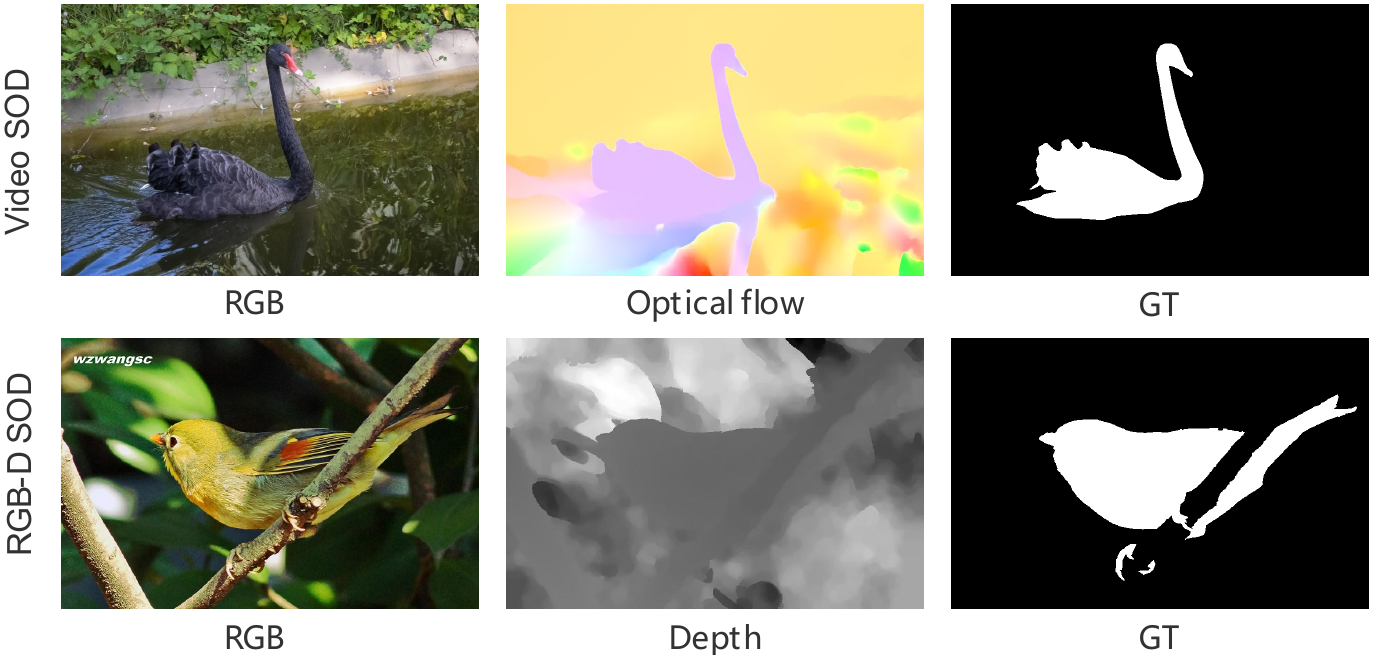,width=0.48\textwidth}}
\caption{Illustration of the common characteristic of optical flow and depth maps, for video and RGB-D SOD.}
\label{fig:Fig_video_SOD_motivation}
\end{figure}
\subsection{On Efficient Video SOD}\label{sec:video_SOD}
Inspired by the encouraging performance of \ourmodel~on RGB-D SOD, in this section, we are motivated to adapt \ourmodel~to the VSOD task to answer the following question: \emph{Does DQFM still work on optical flow (OF) obtained from video data, and can DFM-Net contribute to efficient VSOD?} Note we believe that efficient VSOD is also a promising direction since automatic and efficient video analyses on mobile devices is very useful.

\begin{table*}[htbp]
	\centering
	\caption{Quantitative benchmark results on the VSOD task. The model size is measured by Mb. $\uparrow$/$\downarrow$ for a metric denotes that a larger/smaller value is better. Our results are highlighted in \textbf{bold}. The scores/numbers better than ours are \underline{underlined} (efficient and non-efficient models are labeled separately).}\vspace{-0cm}
	\label{tab:benchmark_vsod}
    \tiny
	\renewcommand{\arraystretch}{1.2}
	\renewcommand{\tabcolsep}{0.6mm}
	\begin{tabular}{lr|cccccccccccccc||ccc}
		\hline\toprule
			& \multirow{4}{*}{Metric} \centering	&DLVS	&MBN	&FGRN	&SCNN	&SCOM	&RSE	&SRP	&MESO	&LTSI	&SSAV	&RCR	&SPD	&FSNet	&\ourmodel*	&PCSA	&Reuse	&\ourmodel	\\
	& 	& \tiny TIP	& \tiny ECCV	&\tiny CVPR	&\tiny TCSVT	&\tiny TIP	& \tiny TCSVT	&\tiny TIP	&\tiny TMM	&\tiny TIP	&\tiny CVPR	&\tiny ICCV	&\tiny TMM	& \tiny ICCV	&Ours	&\tiny AAAI	&\tiny CVPR	&Ours	\\
	&	&2017	&2018	&2018	&2018	&2018	&2019	&2019	&2019	&2019	&2019	&2019	&2020	&2021	&-	&2020	&2021	&-	\\
	&	&\cite{Wang2018VideoSO}	&\cite{li2018unsupervised}	&\cite{li2018flow}	&\cite{tang2018weakly}	&\cite{chen2018scom}	&\cite{cong2019video}	&\cite{cong2019video}	&\cite{xu2019video}	&\cite{chen2019improved}	&\cite{fan2019shifting}	&\cite{Yan2019SemiSupervisedVS}	&\cite{li2019accurate}	&\cite{ji2021FSNet}	&-	&\cite{Gu2020PyramidCS}	&\cite{Park2021Learning}	&-	\\ 
	\midrule
	
\multirow{4}{*}{\begin{sideways}\textit{}\end{sideways}}	&Model size (Mb)	&347.3	&N/A	&N/A	&N/A	&N/A	&N/A	&N/A	&N/A	&132	&247.7	&205.5	&754	&391.4	&\textbf{93}	&19.8	&27.5	&\textbf{8.5}	\\
\midrule  \multirow{4}{*}{\begin{sideways}\textit{DAVIS}\end{sideways}}	&$S_{\alpha}\uparrow$	&.802	&.887	&.838	&.761	&.814	&.748	&.662	&.718	&.876	&.893	&.886	&.783	&\underline{.920}	&\textbf{.915}	&\underline{.899}	&.897	&\textbf{.898}	\\
	&$F_{\beta}^{\rm m}\uparrow$	&.721	&.862	&.783	&.679	&.746	&.698	&.660	&.660	&.850	&.861	&.848	&.763	&\underline{.907}	&\textbf{.898}	&\underline{.880}	&\underline{.884}	&\textbf{.869}	\\
	& $E_{\xi}^{\rm m}\uparrow$     	&.895	&.966	&.917	&.843	&.874	&.878	&.843	&.853	&.957	&.948	&.947	&.892	&.970	&\textbf{.971}	&.965	&.955	&\textbf{.959}	\\
	& $\mathcal{M}\downarrow$ 	&.055	&.031	&.043	&.077	&.055	&.063	&.070	&.070	&.034	&.028	&.027	&.061	&.020	&\textbf{.019}	&\underline{.021}	&\underline{.018}	&\textbf{.025}	\\
\midrule \multirow{4}{*}{\begin{sideways}\textit{FBMS}\end{sideways}}	&$S_{\alpha}\uparrow$	&.794	&.857	&.809	&.794	&.794	&.670	&.648	&.635	&.805	&.879	&.872	&.691	&.890	&\textbf{.899}	&.853	&.744	&\textbf{.889}	\\
	&$F_{\beta}^{\rm m}\uparrow$	&.759	&.816	&.767	&.762	&.797	&.652	&.671	&.618	&.799	&.865	&.859	&.686	&.888	&\textbf{.895}	&.810	&.683	&\textbf{.880}	\\
	& $E_{\xi}^{\rm m}\uparrow$     	&.861	&.892	&.863	&.865	&.873	&.790	&.773	&.767	&.871	&.926	&.905	&.804	&.935	&\textbf{.943}	&.916	&.745	&\textbf{.937}	\\
	& $\mathcal{M}\downarrow$ 	&.091	&.047	&.088	&.095	&.079	&.128	&.134	&.134	&.087	&.040	&.053	&.125	&.041	&\textbf{.031}	&.049	&.074	&\textbf{.034}	\\
\midrule \multirow{4}{*}{\begin{sideways}\textit{MCL}\end{sideways}}	&$S_{\alpha}\uparrow$	&.682	&.755	&.709	&.730	&.569	&.682	&.689	&.477	&.768	&.819	&.820	&.685	&.864	&\textbf{.864}	&.754	&.754	&\textbf{.834}	\\
	&$F_{\beta}^{\rm m}\uparrow$	&.551	&.698	&.625	&.628	&.422	&.576	&.646	&.144	&.667	&.773	&.742	&.601	&\underline{.821}	&\textbf{.810}	&.683	&.679	&\textbf{.768}	\\
	& $E_{\xi}^{\rm m}\uparrow$     	&.810	&.858	&.817	&.828	&.704	&.657	&.812	&.730	&.872	&.889	&.895	&.794	&.924	&\textbf{.935}	&.838	&.836	&\textbf{.906}	\\
	& $\mathcal{M}\downarrow$ 	&.060	&.119	&.044	&.054	&.204	&.073	&.058	&.102	&.044	&.026	&.028	&.069	&.023	&\textbf{.020}	&.038	&.037	&\textbf{.027}	\\
\midrule \multirow{4}{*}{\begin{sideways}\textit{DAVSOD-E}\end{sideways}}	&$S_{\alpha}\uparrow$	&.664	&.646	&.701	&.680	&.603	&.577	&.575	&.549	&.695	&.755	&.741	&.626	&.773	&\textbf{.790}	&.735	&.765	&\textbf{.774}	\\
	&$F_{\beta}^{\rm m}\uparrow$	&.541	&.506	&.589	&.541	&.473	&.417	&.453	&.360	&.585	&.659	&.653	&.500	&.685	&\textbf{.700}	&.647	&\underline{.699}	&\textbf{.684}	\\
	& $E_{\xi}^{\rm m}\uparrow$     	&.737	&.694	&.765	&.745	&.669	&.663	&.655	&.673	&.769	&.806	&.803	&.685	&.825	&\textbf{.838}	&.794	&\underline{.832}	&\textbf{.830}	\\
	& $\mathcal{M}\downarrow$ 	&.129	&.109	&.095	&.127	&.219	&.146	&.146	&.159	&.106	&.084	&.087	&.138	&.072	&\textbf{.066}	&.087	&\underline{.051}	&\textbf{.072}	\\

		\bottomrule
		\hline
	\end{tabular}
	\vspace{-0.cm}
\end{table*}

We first analyze the feasibility of this idea. As shown in Fig.~\ref{fig:Fig_video_SOD_motivation}, the optical flow map plays a complementary role in the VSOD task as the depth map does for RGB-D SOD, \ie{}a depth map provides extra spatial information and an optical flow map provides motion cues. Meanwhile, the optical flow map may also present misleading appearances\eg{broken structures, or erroneous motion}. This similar characteristic between depth and optical flow makes it possible to transfer \ourmodel~to VSOD.  
In addition, we find that the widely used \emph{pretraining strategy} in the VSOD task is less effective for lightweight models, so we propose a simple but effective \emph{joint training strategy}, to help better adapt to VSOD.


 \subsubsection{Joint Training Strategy} 
Most existing optical flow-based VSOD models \cite{ji2021FSNet,fu2021siamese,Gu2020PyramidCS} adopt a pretraining strategy by utilizing extra RGB data for the RGB branch to enhance generalisation, owing to the relatively small size of VSOD datasets. However, almost no work in this field considers if this strategy is suitable for lightweight models. We empirically find that it is less effective for our lightweight model, as will be shown in Tab.~\ref{tab:ablation_train_vsod}. The failure of pretraining for lightweight models has also been reported by \cite{cheng2021highly,poudel2019fast}. 
The underlying reason is that, compared to normal models, lightweight models are easier to converge and also to forget prior information, making such pretraining less influential \cite{cheng2021highly}. 
To alleviate this problem and better leverage extra RGB data, we introduce a strategy called \emph{joint training}. Specifically, we match each RGB image with an all-black image to form a pseudo RGB-flow pair. An all-black image can be seen as an optical flow map for the still case where no motion takes place. Finally, we train the model on the still dataset and VSOD datasets jointly, where the two types of data are mixed up. This proposed strategy improves model generalizability significantly according to Sec.~\ref{sec:ablation_dqfm_vsod}.

 \begin{figure*}[htbp]
  \centering
 \centerline{\epsfig{figure=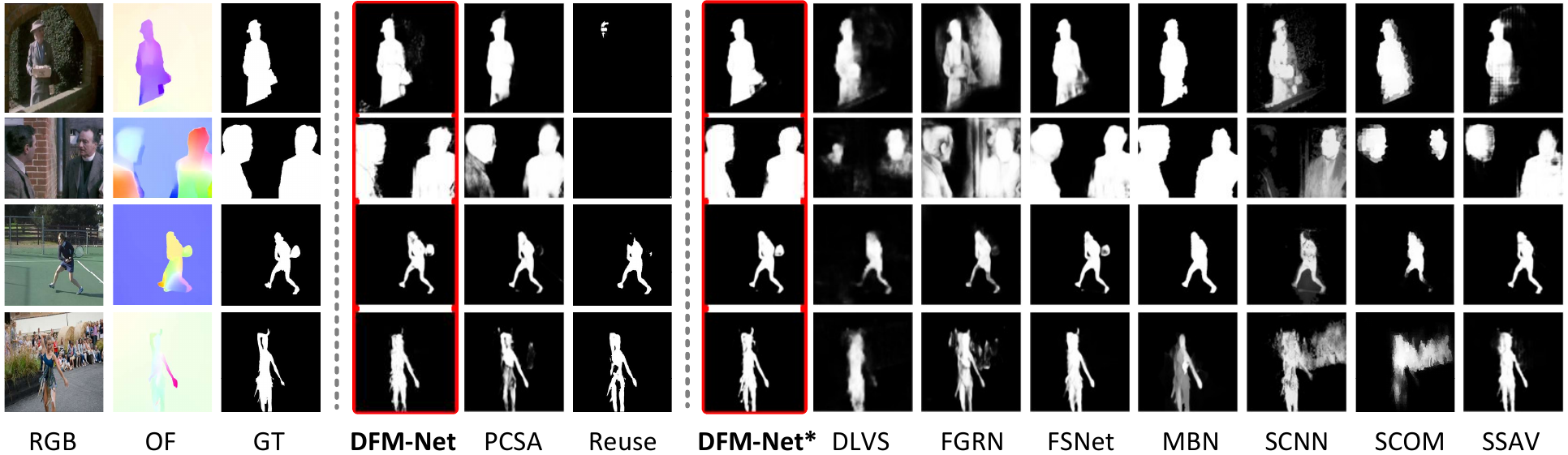,width=1.0\textwidth}}\vspace{-0cm}
\caption{Qualitative comparisons of \ourmodel~and \ourmodel* with SOTA VSOD models. Efficient and non-efficient models are compared separately.}\vspace{-0cm}
\label{fig:visual_results_vosd}
\end{figure*}

\subsubsection{Experimental details}
Following previous works~\cite{ji2021FSNet,fu2021siamese}, we use VSOD datsets DAVIS (30 clips)~\cite{perazzi2016benchmark}, FBMS (29 clips)~\cite{ochs2013segmentation}, and also still RGB SOD dataset DUTS~\cite{wang2017learning} (10,553 images) for training. Tests are conducted on the validation set (20 clips) of DAVIS16, the test set of FBMS (30 clips), the whole MCL~\cite{Kim2015SpatiotemporalSD} (9 clips) as well as the 
easy set of DAVSOD19~\cite{fan2019shifting} (35 clips). 

For other details, we basically follow the settings for RGB-D SOD in Sec~\ref{sec:implementation_details}. For the TDB network, we modify the input channel of the first IRB block from 1 to 3, to accommodate the change from one-channel depth to three-channel optical flow. We obtain optical flow maps by using the recent RAFT~\cite{teed2020raft}.  

\subsubsection{Quantitative and Qualitative Comparisons on VSOD}\label{sec:benchmark_vsod}

\begin{table}[htbp]

    \centering
 	\tiny
	\renewcommand{\arraystretch}{1.3}
	\renewcommand{\tabcolsep}{1.7mm}
	\caption{Efficiency comparison with existing VSOD methods. The model size, $T_{CPU}$, $S_{GPU}$, and $S_{GPU}^*$ are measured by Mb/ms/FPS/FPS, respectively. $S_{GPU}^*$ of Reuse is not available for it only supports one-batch inference. The score better than ours is \underline{underlined}.}\vspace{-0cm}
	\begin{tabular}{c|c|c|ccc}
		\hline\toprule
		
 \multicolumn{2}{c|}{Method} &Model size & $T_{CPU}$ &$S_{GPU}$ &$S_{GPU}^*$ \\\midrule
DLVS \cite{Wang2018VideoSO} &TIP17	&347.3	&4600 &3	&85	\\

RCR \cite{Yan2019SemiSupervisedVS}&ICCV19	&205.5 &975 &18 &25	\\

FSNet \cite{ji2021FSNet}&ICCV21 &391.4 &2600 &25 &75	\\

\rowcolor{mygray}

\ourmodel*~& Ours &\textbf{93} &\textbf{161} &\textbf{62} &\textbf{176}\\
\midrule
PCSA \cite{Gu2020PyramidCS} &AAAI20 &19.8 &176 &\underline{111} &120\\

Reuse \cite{Park2021Learning} &CVPR21	&27.5 &1750  &35 &N/A  \\

\rowcolor{mygray}
\ourmodel~& Ours &\textbf{8.5} &\textbf{53} &\textbf{60} &\textbf{330}\\
		\bottomrule
		\hline
	\end{tabular}
    \label{tab:efficiency_vsod}
	\vspace{-0.2cm}
\end{table}
As shown in Tab.~\ref{tab:benchmark_vsod}, we compare \ourmodel~and \ourmodel*~quantitatively with 15 advanced methods. The accuracy and efficiency metrics follow the criteria in RGB-D SOD experiments. Note that some of these works do not provide code or report the number of parameters, so the model size of these methods are labeled as ``N/A''. As for inference latency, we compare with some representative ones and show the results individually in Tab.~\ref{tab:efficiency_vsod}.

Generally, although \ourmodel~and \ourmodel*~are not specially designed for VSOD (no long-term temporal consideration), they both achieve competitive accuracy while they are much smaller and faster than the contenders. Specifically, \ourmodel~is much more efficient than the recent lightweight models PCSA and Reuse, with 57\% and 69\% parameters decrease. This leads to \ourmodel' s~best performance in terms of $T_{CPU}$ and $S_{GPU}^*$. As for $S_{GPU}$, PCSA runs a little faster than \ourmodel~because PCSA has custom CUDA implementation. Furthermore, according to Tab.~\ref{tab:benchmark_vsod} and Tab.~\ref{tab:efficiency_vsod}, \ourmodel* surpasses almost all the non-efficient competitors in terms of both efficiency and accuracy. 
Qualitative results are exhibited in  Fig.~\ref{fig:visual_results_vosd}, our models generate stable and better predictions under various optical flow qualities.

\subsubsection{Ablation Analyses on VSOD}
\textbf{Effectiveness of DQFM.}\label{sec:ablation_dqfm_vsod}
To answer \emph{whether DQFM still works for optical flow}, we conduct an ablation experiment. As shown in Tab.~\ref{tab:abaltion_dqfm_vsod}, \#13, \#14, \#15 denote the variants of \ourmodel~that are obtained by removing DQFM, DHA, and DQW, respectively, while \#16 indicates the full \ourmodel. Generally, we can see that on the VSOD task, inserting either DQW or DHA, or both, will lead to notable improvements with only little parameter increases. With DQFM equipped, the max F-measure is improved by 6.4\% on DAVSOD. As for the synergy effect of DQW and DHA, it works on almost all datasets except DAVIS.

\begin{table*}[htbp]
	\centering
 	\tiny
	\renewcommand{\arraystretch}{1.3}
	\renewcommand{\tabcolsep}{1.8mm}
	\caption{Ablation analyses for DQFM on four VSOD datasets. The best results are highlighted in \textbf{bold}.}\vspace{-0cm}
	\label{tab:ablation_dqfm_vsod}
	\begin{tabular}{c|c|c|cccc|cccc|cccc|cccc}
		\hline\toprule
		\multirow{2}{*}{\#} 
		&\multirow{2}{*}{DQW}
		&\multirow{2}{*}{DHA}
		
&\multicolumn{4}{c|}{\textbf{DAVIS~\cite{perazzi2016benchmark}}}  &\multicolumn{4}{c|}{\textbf{FBMS~\cite{ochs2013segmentation}}}
		&\multicolumn{4}{c|}{\textbf{MCL~\cite{Kim2015SpatiotemporalSD}}}
		&\multicolumn{4}{c}{\textbf{DAVSOD-E~\cite{fan2019shifting}}}\\ &&
        &$S_{\alpha}$
	    &$F_{\beta}^{\rm m}$
	    & $E_{\xi}^{\rm m}$ 
		&$\mathcal{M}$  
		&$S_{\alpha}$
	    &$F_{\beta}^{\rm m}$
	    & $E_{\xi}^{\rm m}$ 
		&$\mathcal{M}$  
		&$S_{\alpha}$
	    &$F_{\beta}^{\rm m}$
	    & $E_{\xi}^{\rm m}$ 
		&$\mathcal{M}$  
		&$S_{\alpha}$
	    &$F_{\beta}^{\rm m}$
	    & $E_{\xi}^{\rm m}$ 
		&$\mathcal{M}$  \\
		\hline
 
13	&	& 	&.885	&.851	&.949	&.030	&.861	&.862	&.920	&.050	&.814	&.730	&.883	&.032	&.723	&.610	&.781	&.094\\

14	&\checkmark &	 &\textbf{.900}	&\textbf{.875}	&\textbf{.961}	&\textbf{.023}	&.880	&.866	&.925	&.041	&.821	&.741	&.889	&.028	&.771	&.678	&.827	&.073	\\

15	&	&\checkmark 	&\textbf{.900}	&\textbf{.875}	&\textbf{.961}	&.024	&.880	&.871	&.929	&.039	&.829	&.765	&.900	&.027	&.774	&.683	&.827	&.075	\\

\rowcolor{mygray}

16	&\checkmark 	&\checkmark 			&.898	&.869	&.959	&.025	&\textbf{.889}	&\textbf{.880}	&\textbf{.937}	&\textbf{.034}	&\textbf{.834}	&\textbf{.768}	&\textbf{.906}	&\textbf{.027}	&\textbf{.774}	&\textbf{.684}	&\textbf{.830}	&\textbf{.072}\\



         







		\bottomrule
		\hline
	\end{tabular}
	\vspace{-0.2cm}
\end{table*}
\begin{table*}[htbp]
	\centering
 	\tiny
	\renewcommand{\arraystretch}{1.5}
	\renewcommand{\tabcolsep}{1mm}
	\caption{Performance of different training strategies. ``w/o Pretraining'' means direct training on VSOD datasets without DUTS. ``Pretraining'' means using DUTS for pretraining. ``Joint training'' means the proposed joint training strategy.}\vspace{-0cm}
	\label{tab:ablation_train_vsod}
	\begin{tabular}{c|c|c|cccc|cccc|cccc|cccc}
		\hline\toprule
		\multirow{2}{*}{\#} 
		&Training &\multirow{2}{*}{Datasets for training} 
		&\multicolumn{4}{c|}{\textbf{DAVIS~\cite{perazzi2016benchmark}}}  &\multicolumn{4}{c|}{\textbf{FBMS~\cite{ochs2013segmentation}}}
		&\multicolumn{4}{c|}{\textbf{MCL~\cite{Kim2015SpatiotemporalSD}}}
		&\multicolumn{4}{c}{\textbf{DAVSOD-E~\cite{fan2019shifting}}}
\\ &Strategy &
        &$S_{\alpha}$
	    &$F_{\beta}^{\rm m}$
	    & $E_{\xi}^{\rm m}$ 
		&$\mathcal{M}$  
		&$S_{\alpha}$
	    &$F_{\beta}^{\rm m}$
	    & $E_{\xi}^{\rm m}$ 
		&$\mathcal{M}$  
		&$S_{\alpha}$
	    &$F_{\beta}^{\rm m}$
	    & $E_{\xi}^{\rm m}$ 
		&$\mathcal{M}$  
		&$S_{\alpha}$
	    &$F_{\beta}^{\rm m}$
	    & $E_{\xi}^{\rm m}$ 
		&$\mathcal{M}$  
	\\
		\hline
17 &w/o Pretraining &DAVIS+FBMS &.890	&.856	&.959	&.027	&.868	&.860	&.914	&.048	&.830	&.753	&\textbf{.913}	&.029	&.729	&.618	&.799	&.091 \\
18 &Pretraining &DAVIS+FBMS+DUTS &.885	&.853	&.953	&.030	&.879	&.866	&.926	&.042	&.819	&.753	&.885	&.031	&.744	&.637	&.802	&.090 \\
\rowcolor{mygray} 16 &Joint training &DAVIS+FBMS+DUTS	&\textbf{.898}	&\textbf{.869}	&\textbf{.959}	&\textbf{.025}	&\textbf{.889}	&\textbf{.880}	&\textbf{.937}	&\textbf{.034}	&\textbf{.834}	&\textbf{.768}	&.906	&\textbf{.027}	&\textbf{.774}	&\textbf{.684}	&\textbf{.830} &\textbf{.072} \\


		\bottomrule
		\hline
	\end{tabular}
    \label{tab:abaltion_dqfm_vsod}
	\vspace{-0.2cm}
\end{table*}

\textbf{Effectiveness of Joint Training.}\label{sec:ablation_train_vsod} To see \emph{how the pretraining strategy performs for lightweight models,} we conduct several training experiments as shown in Tab.~\ref{tab:ablation_train_vsod}. \zwb{Model \#16 is the proposed \ourmodel~applied to VSOD, which adopts the ``Joint training'' strategy. \#17 means direct training on VSOD datasets DAVIS and FBMS, but without pretraining on DUTS. \#18 is trained on DAVIS and FBMS after pretraining on DUTS for 300 epochs (equal to that for training). }
Comparing between \#17 (w/o Pretraining) and \#18 (with Pretraining) in Tab.~\ref{tab:ablation_train_vsod}, one can see pretraining becomes less effective, with accuracy improvement on FBMS and DAVSOD but decrease on DAVIS and MCL. Moreover, we have experienced double-long training time for \#18. This comparison justifies our motivation to adopt the joint training strategy, resulting in model \#16 to better utilize extra RGB data. As can be seen, \#16 achieves the best results, producing stable and consistent improvements across all datasets, especially on DAVSOD.


\section{Conclusion}\label{sec:conclusion}

In this paper, we introduce an efficient RGB-D SOD model called \ourmodel. It is achieved by equipping a lightweight RGB-D SOD framework with an efficient DQFM process, which can significantly boost the detection accuracy with little time-space consumption increase. DQFM consists of two components namely DQW and DHA.
The DQW weighs the depth features according to the low-level alignment between RGB and depth inputs, while DHA attends to the depth features spatially through multiple holistic attention maps derived from the depth stream and refined with low-level RGB features.
Besides, we also design a tailored depth backbone and a two-stage decoder as the basis of the lightweight framework to further ensure efficiency. Interestingly, the statistics of inferred weights from DQW somewhat reveal its underlying working mechanism, showing that it is feasible to distinguish depth maps of various qualities  without any quality labels.

Extensive experiments on nine RGB-D datasets demonstrate that \ourmodel~breaks records in terms of model size and inference latency while retaining very competitive accuracy for RGB-D SOD. Furthermore, \ourmodel~shows strong performance when applied to the VSOD task and achieves the highest efficiency. We also propose and validate the joint training strategy concerned for lightweight models.
In the future, we believe it would be attractive to deploy \ourmodel~onto embedding/mobile systems that process RGB-D and video data. Not just limited to that, it is worth exploring whether DQFM can be applied to more quality-inspired dense prediction tasks, such as RGB-D and video semantic segmentation, as well as camouflage object detection.

\bibliography{sn-bibliography}


\end{document}